\documentclass[authoryear,preprint,review,12pt]{elsarticle}

\usepackage{amssymb}
\usepackage{amsmath}
\usepackage{booktabs}
\usepackage{siunitx}
\usepackage{array}
\usepackage{rotating}
\usepackage{longtable}
\usepackage{placeins}
\usepackage{url}
\usepackage{tikz}
\usetikzlibrary{arrows.meta, positioning, calc, fit, backgrounds}
\graphicspath{{figures/}}

\newcommand{\tabnote}[1]{\par\smallskip\begin{minipage}{\linewidth}\footnotesize #1\end{minipage}}

\journal{International Journal of Applied Earth Observation and Geoinformation}

\begin{document}

\begin{frontmatter}

\title{Circumpolar Wall-to-Wall Detection of Retrogressive Thaw Slumps with Scarce Labels and Extreme Class Imbalance}

\author[label1]{Yili Yang\corref{cor1}}
\ead{yyang@woodwellclimate.org}
\cortext[cor1]{Corresponding author}
\author[label1]{Heidi Rodenhizer}
\author[label1]{Robb Young}
\author[label1]{Brendan M. Rogers}
\author[label1]{Marlena Bartkus}
\author[label1]{Stefano Potter}
\author[label1]{Greg Fiske}
\author[label1]{Scott Zolkos}
\author[label2]{Wenwen Li}
\author[label2]{Chia-Yu Hsu}
\author[label3]{Luigi Marini}
\author[label3]{Todd C. Nicholson}
\author[label4,label5]{Benjamin C. Maglio}
\author[label1]{Jennifer Watts}
\author[label1]{Anna Liljedahl}
\author[label1]{Susan M. Natali}

\affiliation[label1]{organization={Woodwell Climate Research Center},
            addressline={149 Woods Hole Road},
            city={Falmouth},
            postcode={02540},
            state={MA},
            country={United States}}

\affiliation[label2]{organization={Arizona State University},
            addressline={1151 S. Forest Ave},
            city={Tempe},
            postcode={85287},
            state={AZ},
            country={United States}}

\affiliation[label3]{organization={National Center for Supercomputing Applications (NCSA), University of Illinois Urbana-Champaign},
            addressline={1205 W. Clark St},
            city={Urbana},
            postcode={61801},
            state={IL},
            country={United States}}

\affiliation[label4]{organization={University of Alaska Fairbanks},
            addressline={1731 South Chandalar Drive},
            city={Fairbanks},
            postcode={99775},
            state={AK},
            country={United States}}

\affiliation[label5]{organization={University of Colorado Boulder},
            city={Boulder},
            postcode={80309},
            state={CO},
            country={United States}}
\begin{abstract}
Retrogressive thaw slumps (RTS) are among the most dynamic ground deformations caused by permafrost thaw.
Abrupt thaw processes such as RTS could roughly double the contribution of permafrost thaw to climate warming, and already threaten a third of Arctic infrastructure.
Circumpolar RTS mapping has advanced quickly, yet released inventories together cover only part of the permafrost region.
RTS occupy only 0.1--0.5\% of image tiles and labels are costly, so a detector must be designed and evaluated as a rare-target problem.
We therefore developed a deep learning system and mapped 20.68 M km\textsuperscript{2} of circumpolar permafrost using 41.6 M tiles of 4.8 m PlanetScope imagery.
Training used 22,259 tiles with 2,609 RTS instances in 1,718 of them, and the model was tuned through 172 controlled experiments.
A convolutional network matched every alternative tested, including multimodal fusion, foundation-model encoders and masked-autoencoder pretraining, each at far more compute for similar or worse results.
Labelled data moved performance most, and a foundation model's advantage appeared only when labels were scarce.
Tested once on held-out ecoregions at the real-world prevalence of 0.1--0.5\%, the final model achieved an area under the precision-recall curve of 0.812--0.855, an object precision of 0.854 and F1 of 0.644.
Deployed wall-to-wall, the model produced a circumpolar map of 19,068 high-confidence detections whose sampled precision ranges from 0.54 to 0.90 by size class.
We present an evaluation protocol for rare targets, a controlled account of what governs accuracy under scarce labels, and the circumpolar RTS map the validated model produced.

\end{abstract}


\end{frontmatter}

\section{Introduction}
\label{sec:introduction}

Retrogressive thaw slumps (RTS) are rapidly developing thermokarst landforms formed by abrupt thaw and mass wasting in ice-rich permafrost.
They typically initiate when erosion or another disturbance exposes massive ground ice or ice-rich permafrost on a slope.
The exposed ice then thaws, causing a steep headwall to retreat upslope while thawed sediment and meltwater move downslope, forming a largely unvegetated slump floor or scar \citep{nesterova2024terminology}.
Unlike the gradual deepening of the seasonally thawed layer, an RTS erodes vegetation and soil via ground collapse.
This erosion is sustained by the downslope evacuation of thawed debris, and an active headwall can retreat by more than ten metres in one thaw season \citep{nesterova2024terminology}.
RTS are therefore a clear surface expression of abrupt permafrost thaw, and a class of thermokarst processes that mobilise thawed substrate far faster than gradual warming alone \citep{zolkos2022permafrost, lantuit2005temporal, maier2025masswasting, kokelj2013advances}.
This distinction matters for the carbon cycle.
Abrupt thaw is expected to occur across less than one-fifth of the terrestrial permafrost region.
Yet, given the often rapid exposure of soil carbon and relatively high emissions of CH\textsubscript{4} from the saturated environment, abrupt thaw could roughly double the permafrost contribution to climate warming that gradual thaw alone would produce \citep{turetsky2020abrupt, schuur2022permafrost, schadel2026permafrost}.
RTS also threaten ecosystems, the built environment, and Arctic communities.
About one-third of pan-Arctic infrastructure and nearly half of Russian Arctic hydrocarbon fields stand on ground where permafrost thaw can cause severe damage \citep{hjort2018degrading}.
Downstream propagation of the RTS substrate can alter aquatic food webs and biogeochemical cycles across watershed scales, potentially increasing toxic mercury levels and chronically stressing basal food web organisms \citep{pierre2018unprecedented, levenstein2021effects, kokelj2021thaw}.
Mapping where RTS occur is thus a necessary step towards constraining the wider permafrost carbon feedback \citep{schuur2015climate, turetsky2020abrupt} and anticipating risks to ecosystems, infrastructure, and communities.

RTS activity has increased in some regions as the Arctic warms and becomes wetter \citep{kokelj2015increased, mccrystall2021new}.
Summers of extreme heat have triggered thousands of new slope failures in a single season, and repeat imaging shows rising numbers and area of RTS across several Arctic regions \citep{lewkowicz2019extremes, barth2025rapid}.
Recent hemispheric work has begun to measure not only where RTS occur but how much material they move, reporting large and accelerating mass loss over the past decade \citep{dai2025volumetric, maier2025masswasting}.
Comparable acceleration has been documented beyond the Arctic on the Qinghai-Tibet Plateau \citep{xia2024widespread}.
These findings underscore the need for continuous detection of RTS against which such change can be measured \citep{nitze2025darts}.

Automated RTS mapping using deep learning (DL) has advanced quickly to meet this need.
Early studies trained convolutional networks to segment RTS from high-resolution optical imagery within single regions \citep{huang2020using, yang2023mapping, nitze2021testing, witharana2022automated}.
Later work increased model complexity, fusing optical imagery with elevation and multi-temporal trend layers and adopting transformer and foundation-model backbones \citep{li2025multimodal, zeng2026multisource, yuan2025amrtsnet, nitze2025darts}.
A parallel line of work detects RTS from change in satellite image time series rather than from a single image, measuring their growth from multi-temporal imagery and their volume from differential elevation models \citep{huang2021beiluhe, ma2025imagestacks, maier2025elevation}.
The most recent efforts extend towards continental and hemispheric coverage \citep{nitze2025darts, lou2026hemispheric}.
This progress rests on a shared labelling effort, now consolidated in the Arctic Retrogressive Thaw Slumps (ARTS) compilation, which brings together tens of thousands of digitised RTS and background features under a common standard \citep{yang2025arts}.

This progress leaves three problems open for a circumpolar detector.
First, the coverage of RTS maps across the Northern Hemisphere permafrost region is still partial and sporadic.
Second, reported accuracy is often measured internally, and model performance assessed on held-out data from the same regions tends to exceed agreement with independent reference datasets \citep{nitze2025darts}.
Part of this gap reflects evaluation design.
When training and test samples are drawn from neighbouring ground, spatial autocorrelation inflates apparent accuracy, a pitfall now well documented for both ecological mapping and convolutional segmentation \citep{roberts2017cross, ploton2020spatial, kattenborn2022spatially, meyer2022global}.
Third, RTS are, overall, rare on the landscape, occupying well under 1\% of image tiles.
Yet, detection is usually reported with metrics that are blind to extreme class imbalance, and therefore overstate the precision achievable in operational mapping \citep{saito2015precision}.
This rarity carries an operational cost that earlier systems make plain.
When a detector scans millions of tiles, false detections outnumber true ones even at a low false-positive rate.
The two errors are also unequal in consequence, because false positives destroy map integrity and skew statistical results.
False negatives leave a map less complete but keep its precision high, yielding a conservative but completely reliable map.
The largest automated product counters this with a rich engineered input, stacking the normalised difference vegetation index (NDVI), multi-decadal tasselled-cap trends, relative elevation and slope onto the spectral bands \citep{nitze2025darts}.
A usable map is then obtained by screening the output against external datasets, particularly those including thaw-like features such as water bodies and human impacts \citep{nitze2025darts}.
The residual false positives are thus managed after detection with ancillary data rather than reduced at the point of detection, which points to the design of the mapping system, not the capacity of the network, as the binding constraint.
Underlying all three is the cost of labels.
Even expert annotators disagree substantially when outlining the same RTS on remote sensing images, which limits how far exhaustive manual labelling can scale \citep{nitze2025labeling}.

This study sets out to meet all three at once across the circumpolar region.
Addressing this raises a question of interest well beyond permafrost science.
In a setting defined by scarce labels, extreme class imbalance and small, rare targets, does the recent turn towards remote-sensing foundation models and heavier multimodal fusion actually improve detection, or do model capacity and input representation trade off against the one resource these studies hold fixed, the labelled data itself?
The question is timely because task-specific networks have been shown to match or exceed fine-tuned geospatial foundation models when labels are limited \citep{marsocci2026pangaea}, and RTS mapping is an unusually clean test case for foundation-model encoders and learned satellite embeddings \citep{simeoni2025dinov3, ravi2024sam2, brown2025alphaearth}.

We address both the practical and methodological questions with a single Earth Observation system that maps RTS across the circumpolar permafrost region using PlanetScope quarterly basemap imagery, treating each stage of the system as a design choice to be tested.
Our contributions are threefold.
First, we deploy the validated model across the 20.68 million km\textsuperscript{2} circumpolar domain of 41.6 million image tiles, producing a circumpolar map of high-confidence RTS detections obtained without ancillary-data masking, whose precision we characterise from a stratified sample.
Second, we adopt an evaluation protocol tailored to this regime, combining ecoregion-blocked spatial splits, prevalence-scaled precision-recall metrics, and a single-shot test on held-out regions \citep{roberts2017cross, kattenborn2022spatially, saito2015precision}.
Third, we compare training data quantity, model capacity, self-supervised pretraining, and input representation as competing routes to better detection, varying each under a fixed recipe and a fixed adoption criterion.
Earlier RTS studies have compared architectures or input channels, but none has varied all four factors under one criterion.

\section{Study area and data}
\label{sec:data}

\subsection{The circumpolar inference domain}
\label{sec:domain}
The study covers the circumpolar permafrost region of the Northern Hemisphere up to approximately 76\,\textdegree{}N.
We defined a mapping domain as the intersection of the Arctic-boreal region with the permafrost region \citep{obu2019northern}, where the permafrost region includes any area that holds some permafrost within a 100 by 100 km cell.
The limiting layer on this domain is the coverage of the PlanetScope quarterly basemap, which is guaranteed only to 74\,\textdegree{}N and reaches about 76\,\textdegree{}N in most places.
The part of the domain covered by PlanetScope imagery forms the inference domain of this study.
This domain measures 20.68 million km\textsuperscript{2}, compared to 21.34 million km\textsuperscript{2} for the parent region under the same 100 by 100 km rule, and extends from about 45\,\textdegree{}N at the boreal margin to about 76\,\textdegree{}N.
The rule is deliberately generous, so at the boreal margin the domain reaches beyond the permafrost zonation classes of \citet{obu2019northern}.
Section~\ref{sec:disc-comparison} therefore measures coverage against that zonation when comparing inventories.
The rest of the permafrost region lies beyond the basemap and is not mapped here.

We used the ecoregions of \citet{dinerstein2017ecoregion} as spatial units, keeping those in the tundra and boreal forest biomes and clipping them to the domain.
This gives 49 ecoregions, mapped over the inference domain in Supplementary Section~S6, which serve as the blocking units of the training splits defined in Section~\ref{sec:splits}.

\subsection{Satellite imagery}
\label{sec:imagery}
The primary imagery is the PlanetScope Global Quarterly Basemap \citep{planet2025basemaps}, a cloud-free mosaic composited over the July to September growing season.
We used the 2024 basemap for training and the 2025 basemap for deployment.
The basemap is a visual product prepared for computer-vision analytics rather than a radiometric measurement.
The provider composites, colour-corrects and rescales the source scenes to 8-bit values through a proprietary harmonisation, so the pixel values are display-ready rather than surface reflectance.

The basemap provides three visible bands at a projected resolution of 4.77\,m in Web Mercator at zoom level 15.
The ground sample scales with latitude as $4.77 \cos(\text{lat})$\,m, which gives about 1.63\,m at 70\,\textdegree{}N.
The effective resolution is coarser than either figure, near 10\,m due to the field of view of the sensor rather than a processing step \citep{kim2022system}.
All tile dimensions quoted below are in projected units, so a 512 by 512 training tile is about 2.4\,km on a side in projection and correspondingly smaller on the ground at high latitude.
The 2025 imagery is delivered as 4096 by 4096 pixel RGBA quads on the same grid, where the alpha band marks no-data, and a full domain holds 309,100 quads.
Image tiles at inference are read as 512 by 512 windows that may cross quad boundaries.

Four further collections supply the candidate auxiliary channels: Sentinel-2 surface reflectance from the harmonised Level-2A collection, the Google Satellite Embedding dataset \citep{brown2025alphaearth}, ArcticDEM V4.1 \citep{porter2023arcticdem} terrain derivatives, and Sentinel-1 radar backscatter.
Section~\ref{sec:channels} describes how these were processed into candidate channels.

\subsection{Reference RTS inventory}
\label{sec:artsdata}
Reference polygons were seeded from the Arctic Retrogressive Thaw Slumps (ARTS) compilation \citep{yang2025arts}.
ARTS supplies confirmed RTS locations across the circumpolar region and, equally important for a rare-target detector, confirmed non-RTS locations from which the negative examples are drawn.
It does not supply RTS outlines digitised on the 2024 basemap used here, so every feature was refined against that imagery under the labelling protocol of Section~\ref{sec:labels}.

\section{Methods}
\label{sec:methods}\label{sec:overview}
\begin{figure}[!htbp]
\centering
\includegraphics[width=\linewidth]{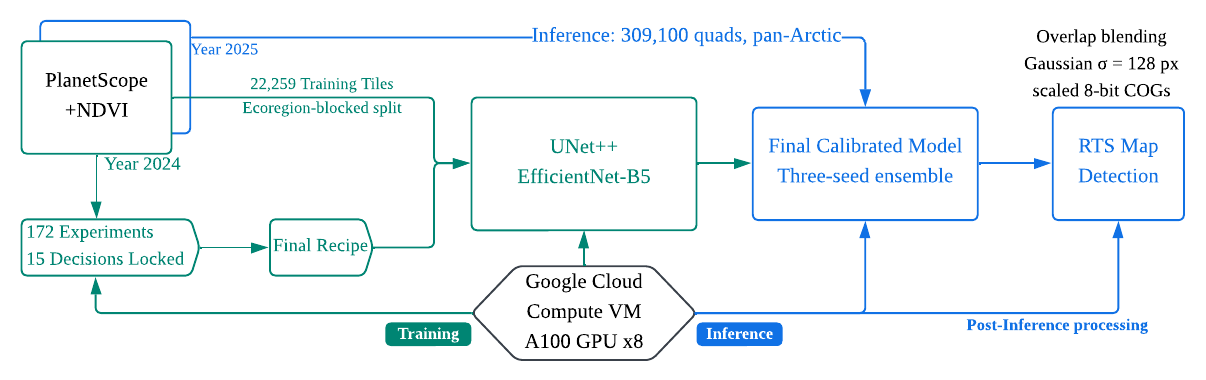}
\caption{End-to-end system overview.
Training drew 22,259 tiles from the 2024 PlanetScope basemap and its Sentinel-2 NDVI channel, split them by ecoregion, and locked the recipe through 172 controlled experiments (Section~\ref{sec:experiments} and Supplementary Tables~S5 and~S6) before training the model that becomes the final calibrated three-seed ensemble.
Deployment applies that ensemble to the 309,100 PlanetScope quads of the 2025 basemap over the pan-Arctic domain.
Both stages run on a Google Cloud virtual machine with eight A100 GPUs.
Post-inference processing blends overlapping tile predictions by an edge-zeroed Gaussian weighting ($\sigma = 128$\,px), writes the calibrated probability field as scaled 8-bit cloud-optimised GeoTIFFs, and vectorises it into the detection map (Section~\ref{sec:postprocessing}).}
\label{fig:system}
\end{figure}

This section describes the end-to-end methods, from data preparation to map production.
Figure~\ref{fig:system} gives an overview of the DL system as two major stages, training and deployment.
The training stage builds a labelled tile dataset from the 2024 basemap, then fixes a model and a training recipe through a programme of controlled experiments.
By recipe, we mean the full set of adopted training choices, from architecture and loss to augmentation and sampling.
The deployment stage applies the final model to the 2025 basemap over the whole inference domain and converts its probability field into a detection map.
The subsections below describe the implementations in order.

\subsection{Label refinement and tiling}
\label{sec:labels}
We manually refined or redrew 2,069 RTS polygons of Section~\ref{sec:artsdata} on the 2024 PlanetScope imagery, producing 2,609 individual polygons that were tiled into 1,718 positive training tiles.
Positive refers to RTS presence in the training tile.
We also generated 20,541 negative tiles in which no RTS is present.
A feature was labelled as an RTS only when three conditions were confirmed: a visible headwall with a cast shadow, a barren slump floor, and a clear morphological contrast with the surrounding terrain.
Features that failed any of the three conditions were excluded.
These excluded features comprised those too small to show the diagnostic traits at basemap resolution, ambiguous features without a clear headwall shadow, stabilised RTS without a barren floor, and debris tongues distant from the slump floor.

The conditions are conjunctive, so all three must hold together.
The shadowed headwall marks the concave scar of collapse, the barren floor marks ongoing thaw, and the morphological contrast separates the RTS from undisturbed ground.
Each trait alone is common in Arctic imagery, where many disturbances expose bare ground or cast sharp shadows, but their joint occurrence is distinctive.
Requiring all three loses marginal features and keeps the positive labels clean.
A looser standard would teach the model that bare ground alone is an RTS, across a domain where bare ground is common and RTS are not.
Label boundaries follow the visible colour and texture contrast between disturbed and undisturbed ground.
Where no contrast exists, the endpoints of the headwall arc are joined with a natural curve so that every outline reads as a single collapsed landform.
An explicit standard also disciplines the human side of labelling.
Four labellers produced the dataset, and because inclusion rests on stated diagnostic conditions rather than individual judgement, their decisions remained consistent.
As a final safeguard, the first author checked every label in a closing review-edit pass, so at least two labellers examined each polygon.

Labels take three values: background=0, RTS=1, and ignore=255.
Fixed tiling cuts some RTS in half, and a fragment showing only a floor or only a headwall cannot be confirmed from its own tile.
Calling it RTS would teach the model that a bare floor is enough, and calling it background would punish the model for finding a real RTS.
Such pixels are marked ignore and dropped from the loss.
Every tile is judged on its own, with neighbouring tiles masked, because inference sees one tile at a time.
The same value is applied to RTS-like features that cannot be confirmed due to basemap quality.
Ignoring a partial feature in one tile need not cost its detection, because the inference stride is set so that 99.5\% of RTS fall completely within at least one tile (Section~\ref{sec:inference}).
The training dataset holds 22,259 tiles, of which 1,718 contain RTS and 20,541 do not.
All negative tiles are drawn from confirmed non-RTS locations in the ARTS compilation.

\subsection{Input channels}
\label{sec:channels}
The final model reads the visible RGB bands and a single auxiliary channel, NDVI, derived from the Sentinel-2 surface reflectance of Section~\ref{sec:imagery}.
We built a growing-season composite from the harmonised Level-2A collection, masking clouds with the scene quality band, keeping scenes below 20\% cloud cover, and taking the per-pixel median over 1 July to 30 September.
NDVI is the normalised difference of the near-infrared and red bands, which the composite provides at 10\,m in EPSG:3857.
The 10\,m NDVI is brought onto the 4.77\,m RGB grid by bilinear resampling in the shared Web Mercator projection and co-registered with the RGB tile through its bounding box, so the two inputs form one aligned 512 by 512 stack.

\begin{figure}[!htbp]
\centering
\includegraphics[width=\linewidth]{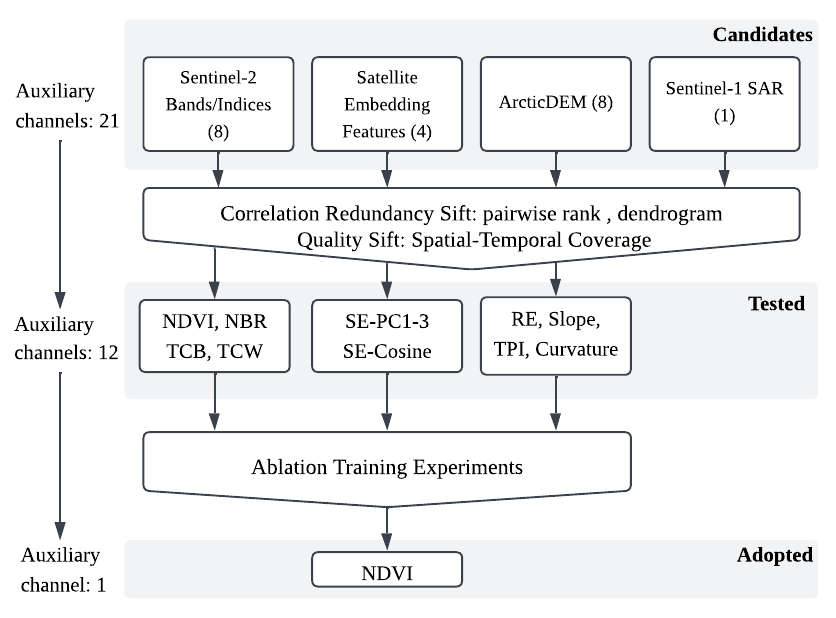}
\caption{The three-stage channel selection.
Candidate auxiliary channels were assembled from four sources: Sentinel-2 bands and indices, Google Satellite Embedding features, ArcticDEM terrain derivatives, and Sentinel-1 radar backscatter.
Two screens were applied to the 21 candidates.
A redundancy screen compared every pair of spectral, embedding and visible layers by pairwise rank correlation and clustering, and a quality screen judged each source on its spatial and temporal coverage.
Sentinel-1 was left at the quality screen for sparse coverage and no distinct RTS signal.
Twelve channels survived, the eight-band auxiliary stack of Supplementary Table~S7 and four ArcticDEM terrain derivatives carried into a separate evaluation (Supplementary Section~S3), namely relative elevation (RE), slope, the topographic position index (TPI), and curvature.
A controlled training ablation (Section~\ref{sec:ablations}) reduced these twelve to one.
NDVI is the only auxiliary channel adopted, and the deployed model reads it alongside the PlanetScope visible bands.}
\label{fig:chanselect}
\end{figure}

NDVI was not chosen in isolation.
It was the single survivor of a three-stage selection that began with 21 candidate auxiliary channels drawn from four sources, narrowed them by removing highly correlated ones, and then let a training ablation decide what actually improved detection.
The effectiveness of NDVI in detecting RTS was also verified in previous studies \citep{li2025multimodal}.
Figure~\ref{fig:chanselect} sketches the three stages, which the rest of this subsection describes and Supplementary Section~S2 documents in full.

The candidate pool was drawn from four data sources chosen to represent the physical signatures of an active RTS in complementary ways (Figure~\ref{fig:chanselect}).
They are Sentinel-1, Sentinel-2, Google's Satellite Embedding dataset (SE) and the ArcticDEM.
Sentinel-2 supplies NDVI, the normalised burn ratio (NBR), the normalised difference water and moisture indices (NDWI and NDMI), and the tasselled-cap brightness and wetness (TCB, TCW) components with the coefficients of \citet{shi2019tasseled}, along with the underlying near-infrared and shortwave-infrared bands.
Sentinel-1 SAR could contribute a physical signal the optical sensors do not carry, namely surface roughness.

The SE dataset \citep{brown2025alphaearth} contributes a learned representation of landscape context.
For every 10 m on Earth and every year from 2017 to 2025, the foundation model AlphaEarth \citep{brown2025alphaearth} learned from petabytes of raw sensor data (optical, radar, structural, climate, topographic, and geotagged text) to represent that location as a single 64-dimensional `fingerprint'.
The embedding is therefore a compressed representation of many years of Earth observation at one place.
We expect it to carry temporal and multi-sensor context that a single physical sensor cannot see.
We used two dimensionality reduction methods to bring the 64-dimensional embedding down to a width suitable as model input.

The first method takes the first three axes of a principal component analysis (PCA) fitted to a circumpolar sample of the SE dataset (Figure~\ref{fig:chanselect} SE-PC1-3, Supplementary Figure~S5).
They are not physically interpretable quantities but a learned representation on a 64-dimensional unit sphere latent space.
The second method (Figure~\ref{fig:chanselect} SE-Cosine, Supplementary Figure~S5) takes the cosine similarity between each pixel embedding and an RTS prototype, the unit-normalised mean embedding of labelled slump pixels from the positive training tiles.
The cosine between two embeddings encodes their similarity, so two pixels with close embeddings describe similar environments however far apart they lie.
This channel therefore rates every pixel by its similarity to the average known RTS pixel.

ArcticDEM terrain derivatives and Sentinel-1 radar backscatter completed the pool as candidates that could, in principle, resolve the concave headwall geometry and the surface roughness of disturbed ground.
The terrain candidates were relative elevation, slope, the topographic position index, and curvature, each of which measures local shape rather than absolute height.

We reduced this pool of 21 channels before training a model to ensure each channel introduced unique and useful information about the presence of RTS.
Sentinel-1 SAR was removed first due to poor coverage and signal quality.
Its usable coverage over the inference domain is sparse, and its backscatter is noisy and shows no clear visible signature that separates an RTS from its surroundings.
The layers were then screened for correlation redundancy.
Each pair of candidate channels was compared by Spearman rank correlation (Supplementary Figure~S2) and a dendrogram (Supplementary Figure~S3) over a held-out tile sample, in three pixel regimes weighted towards the near-boundary pixels where the model earns its score, and a candidate was dropped when it duplicated a retained channel.
This removed the strongly collinear indices, most clearly the normalised difference water index, whose rank correlation with NDVI reached 0.99.
The screen left an eight-band auxiliary stack of NDVI, NBR, the tasselled-cap brightness and wetness pair, the three embedding principal components, and the cosine similarity, listed with its sources in Supplementary Table~S7.
Each of these eight carries information distinct enough from the others to be worth a training run.

Elevation derived from the ArcticDEM provides an index of terrain, in contrast to spectral metrics from the satellite imagery.
We selected terrain derivatives on geomorphological grounds and then tested them.
Seven elevation derivatives were rendered and inspected over a sample of labelled RTS: relative elevation, hillshade, curvature, shaded relief, slope, aspect, and the topographic position index.
Four were carried forward as a separate auxiliary stack used only for the terrain runs, namely relative elevation over a 500\,m neighbourhood, slope, the topographic position index over 300\,m, and curvature.
Hillshade, shaded relief and aspect were excluded as renderings of illumination and slope orientation that duplicate slope itself.

Two properties of the ArcticDEM shaped the experiment before any run.
The first is the age of the elevation surface.
The ArcticDEM mosaic (2007--2021) predates the labels everywhere, and therefore it cannot contain an RTS that developed after it was built.
In addition, the terrain under every tile is several years older than the feature it is asked to describe, resulting in spatial mismatches.
The second is that the ArcticDEM does not cover our entire domain of interest.
ArcticDEM covers every positive tile, but not every negative one, and an absent elevation surface normalises to a constant.
Every ArcticDEM experiment was therefore restricted to the covered tiles, keeping all positives, which makes these scores comparable only with each other.
Supplementary Section~S3 gives the acquisition dates and the coverage figures behind both properties.

The four retained derivatives were then tested together in eight runs against controls trained on the same restricted tiles, under the experiment design of Section~\ref{sec:experiments}.
Section~\ref{sec:ablations} reports the outcome, and Supplementary Section~S3 gives the full results with their caveats.

The third and final stage was a training ablation, which added each of the eight channels to the RGB bands one at a time and then tested the surviving combinations, under the experiment design of Section~\ref{sec:experiments}.
Section~\ref{sec:ablations} reports its outcome.

\subsection{Ecoregion-blocked splits}
\label{sec:splits}
We split the data by ecoregion \citep{dinerstein2017ecoregion} so that whole regions, not individual tiles, are assigned to the training, validation, or test set.
RTS in nearby tiles share geology, permafrost conditions, vegetation and acquisition conditions, so a random tile-level split would test the model on ground effectively adjacent to its training tiles and overstate generalisation.
We targeted an 80/10/10 split of tiles under three constraints: the test set must hold at least 100 positive tiles for a stable precision-recall estimate at 1:1000 prevalence, the validation set must span at least two ecoregions, and the training set should hold at least 80\% of all positive tiles.

\begin{figure}[!htbp]
\centering
\includegraphics[width=\linewidth]{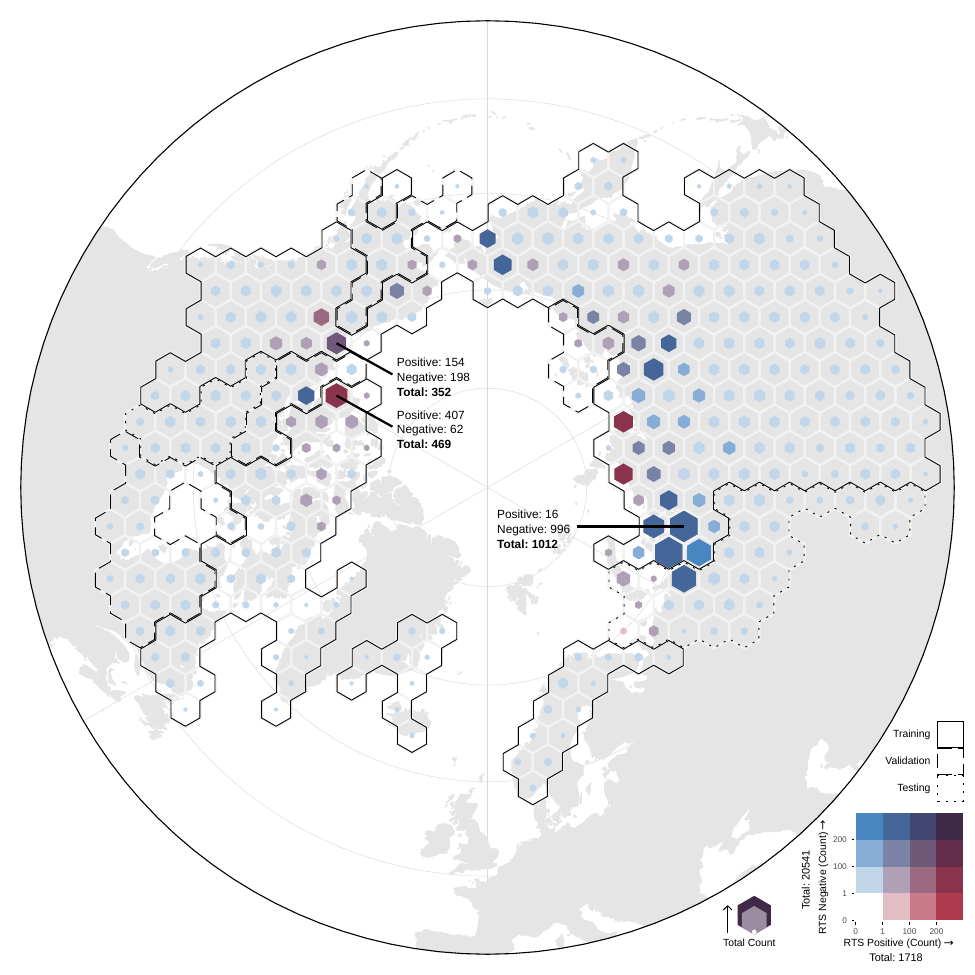}
\caption{Distribution of the training labels across the inference domain.
Each hexagonal cell is shaded by a bivariate scale for the number of positive and negative training tiles it contains, and the solid, dashed, and dotted outlines mark the training, validation, and test ecoregion blocks.
Call-outs give the positive and negative tile counts of selected cells.}
\label{fig:trainmap}
\end{figure}

The split was then verified on the ground: no ecoregion spans more than one set, and a projected-space check found no overlapping tile pairs across splits, with the tightest gap between a test and a training tile just above one tile edge.
The final split assigns 17,951 tiles, 1,513 of them positive, to 40 training ecoregions, 2,151 tiles with 98 positives to five validation ecoregions, and 2,157 tiles with 107 positives to four test ecoregions.
Figure~\ref{fig:trainmap} maps this distribution.

\subsection{Model architecture and training}
\label{sec:architecture}
We framed RTS mapping as a binary semantic segmentation problem.
The final model is a UNet++ decoder \citep{zhou2018unetpp} on an EfficientNet-B5 encoder \citep{tan2019efficientnet} pretrained on ImageNet \citep{deng2009imagenet, russakovsky2015imagenet}.
The input is a four-channel stack of the PlanetScope RGB bands and NDVI, fused at the input layer via channel stacking.
We compared this architecture against a wide set of alternatives, including the FPN \citep{lin2017fpn}, DeepLabV3+ \citep{chen2018deeplabv3plus}, PSPNet \citep{zhao2017pspnet}, and MAnet decoders \citep{fan2020manet}, other EfficientNet sizes, and transformer and foundation-model encoders such as DINOv3 \citep{simeoni2025dinov3} and SAM2 \citep{ravi2024sam2}.
Section~\ref{sec:levers} reports the outcome of these comparisons.

We trained with the AdamW optimiser \citep{loshchilov2019adamw} at a weight decay of $10^{-2}$ and gradient clipping at a norm of 1.0.
Training ran in two phases.
In the first phase, the encoder was frozen for ten epochs while the decoder trained at a constant learning rate of $3\times10^{-3}$.
In the second phase, the whole network was fine-tuned.
The learning rate rose linearly over five epochs to a base of $10^{-3}$ and then followed a single cosine decay to a floor of $10^{-6}$, which is the annealing schedule of \citet{loshchilov2017sgdr} without its warm restarts.
The encoder was held at one tenth of the decoder learning rate throughout the second phase.
Training used a batch size of 32 in BF16 mixed precision \citep{micikevicius2018mixed} and stopped early on a smoothed validation score with a patience of five evaluations.

The loss was a focal loss \citep{lin2017focal} with $\gamma = 2$ and $\alpha = 0.25$.
To keep uncertain object edges from dominating the gradient, we ignored a two-pixel margin around each label boundary.
We maintained an exponential moving average of the weights with a decay of 0.999.
We used it for validation and deployment, a form of weight averaging that widens the optimum and improves generalisation \citep{izmailov2018averaging}.
Because RTS are rare, each training batch was drawn with balanced sampling at a positive fraction of one-half.
Augmentation combined geometric transforms with TrivialAugment \citep{muller2021trivialaugment}, from which the random-scale transform was removed.
Image channels were standardised by a per-dataset z-score, with statistics computed on the training split alone by a one-pass Welford method \citep{welford1962note} and clipped at the 0.1 and 99.9 percentiles.
We trained three models with seeds 42, 43, and 44.

\subsection{Experiment design and adoption criterion}
\label{sec:experiments}
The metric was the precision-recall area under the curve (PR-AUC) to handle the extremely imbalanced class ratio \citep{saito2015precision}, computed as a geometric mean over the prevalence ratios 1:5, 1:10, and 1:20.
Every candidate change was judged against a fixed adoption criterion over the course of three runs.
A change was adopted only when
\begin{equation}
\bar{\Delta} \;\ge\; G = \max(2\sigma_0,\, 0.01) = 0.0112
\quad\text{and}\quad
\Delta_s > 0 \;\; \text{for all } s,
\label{eq:gate}
\end{equation}
with $\Delta_s$ the change in PR-AUC of seed $s$, $\bar{\Delta}$ its three-seed mean, $G$ the required gain, and $\sigma_0 = 0.0056$ the seed standard deviation of the baseline.
The criterion guards against changes that only appear to help because of random variance.
It decides whether a candidate is worth adopting against the recipe already in place, so it cannot separate two candidates that both meet it.
Where two changes were indistinguishable under the criterion, we kept the simpler or less tuned of the two.
Section~\ref{sec:ablations} reports the two decisions that rest on this second rule.

The 172 runs were organised into ten groups that tested normalisation, data volume, loss and boundary handling, input channels and fusion, encoders and foundation models, augmentation, sampling, multi-scale capability, the training data version, and a three-seed confirmation of the final recipe.
Beyond the simple channel stacking of Section~\ref{sec:channels}, the channel group also tested four fusion architectures for the auxiliary bands, each described in Supplementary Section~S1.
Temperature scaling \citep{guo2017calibration} and ensembling were then applied to the final recipe on the validation set rather than as separate training runs.

The multi-scale experiments tested whether a wider field of view could recover large RTS that exceed the 2.4 km footprint of a native-resolution tile.
The training tiles were extended with a half-resolution re-stage of the same scenes, in which 1024 by 1024 windows read at the native 4.77 m projected resolution were bilinearly downsampled to the standard 512 by 512 input, giving an effective 9.55 m per pixel.
The two resolutions were then trained jointly.
Labels were revised accordingly.
Models were evaluated at native resolution to confirm that the deployed configuration was not degraded, and at half resolution to test whether the model had adapted to the coarser scale.
The inference pipeline can likewise run each tile at several scales and fuse the probabilities by a per-pixel mean.
Adoption was judged against three criteria fixed in advance, requiring no regression at native resolution, a usable half-resolution capability, and a gain from fusing the two scales at inference.
Section~\ref{sec:pretraining} reports the outcome.

A follow-on self-supervised masked-autoencoder pretraining tested whether domain adaptation could further improve the strongest foundation encoder.
We continued pretraining the satellite DINOv3 ViT-L encoder with a masked-autoencoder objective \citep{he2022mae} on 295,429 unlabelled tiles from the inference domain, represented as four-channel RGB and NDVI stacks at 512-pixel input size, with every tile that intersects a validation or test region excluded.
Three-quarters of the input patches were masked and replaced by a learnable token, and the loss was the mean squared error on the normalised pixels of the masked patches.
Pretraining ran for 80 epochs on eight A100 GPUs.
The adapted encoder was then fine-tuned under the final recipe with three seeds and evaluated in a three-way comparison, with the fixed order, against the EfficientNet-B5 baseline and the unadapted satellite foundation model under the same recipe.
Section~\ref{sec:pretraining} reports the outcome.

\subsection{Accuracy metrics}
\label{sec:evaluation}
We report both pixel and object metrics.
At the pixel level, we use intersection over union (IoU) and the F1 score for the RTS class.
At the object level, the thresholded map is split into connected components, each scored by its mean interior probability, and these are matched to the reference polygons greedily in descending order of confidence: each detection claims the best-overlapping reference not already taken, and the pair is accepted as a true positive at an IoU of at least 0.3.
Unmatched detections are false positives and unmatched references false negatives, from which object precision, recall, and F1 follow.

Because the true prevalence of RTS is very low, near 0.1 to 0.5\% of tiles, we report the PR-AUC at prevalence ratios of 1:200, 1:500, and 1:1000.
These are computed from a single inference run by subsampling the negatives to each ratio.
By detection settings we mean the two values that turn the probability field into detections, the selection threshold and the minimum object size.
Object metrics are reported at a minimum detected-object size of 80 pixels, the size fixed with the rest of the detection settings before the test set was scored.
Finally, we report the undetectable fraction, the share of reference objects whose footprint holds no predicted probability above 0.30, which bounds the recall any threshold could reach.

We report the calibration gap, the difference between the mean probability the model states for a set of pixels and the frequency with which those pixels are in fact RTS, which measures whether a model-output probability matches the real-world likelihood.

\subsection{Circumpolar inference}
\label{sec:inference}
\begin{figure}[!htbp]
\centering
\includegraphics[width=\linewidth]{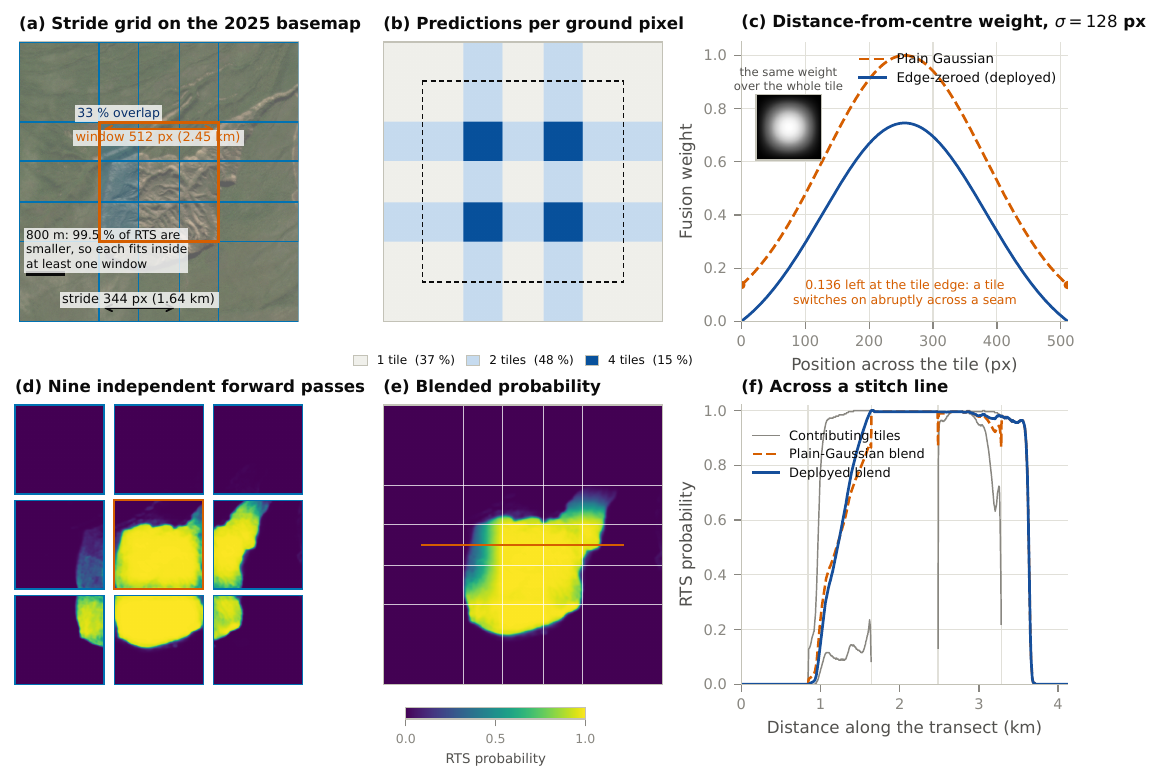}
\caption{Tiling, blending and stitching at inference, shown over one
3 by 3 block of the deployed grid from the delivered 2025 Q3 run
(67.6\,\textdegree{}N, 134.8\,\textdegree{}E).
(a) The stride grid on the 2025 basemap, with one window outlined and the
overlap with its neighbour shaded. The bar gives the 800\,m projected RTS
extent that the stride is chosen to accommodate.
(b) The number of tiles predicting each ground pixel. Within the dashed
interior, where the block reproduces the coverage of the full grid, 37\% of
pixels are predicted once, 48\% twice and 15\% four times.
(c) The fusion weight across a tile. The plain Gaussian retains 0.136 of its
weight at the tile edge, so a tile enters the average abruptly at a seam, while
the deployed weight falls to zero there.
(d) The nine overlapping windows of the block, each one independent forward pass. The
RTS appears in six of them, at a different position in each.
(e) The blended field, with the tile edges drawn and the transect of panel (f)
in orange.
(f) Probability along that transect: each contributing tile, the deployed
blend, and the plain-Gaussian blend.}
\label{fig:tiling}
\end{figure}

The final three-seed ensemble was run over the full domain for the year 2025.
Tiles were read at a stride of 344 pixels, which gives about 33\% overlap.
This stride is chosen from the size distribution of about 40,000 RTS polygons in the ARTS dataset, so that at least one tile fully contains any RTS up to about 800\,m across in projected units, which covers 99.5\% of the population (Figure~\ref{fig:tiling}a).
Every tile is predicted independently, and the overlap gives most ground pixels more than one prediction (Figure~\ref{fig:tiling}b, d).
Overlapping predictions are combined by a distance-from-centre weighted mean, with a Gaussian weight ($\sigma = 128$ pixels) zeroed at the tile edge (Figure~\ref{fig:tiling}c, f).

The network emits a score between 0 and 1 for every pixel, but nothing in training forces that score to behave as a frequency.
Ideally, of all the pixels the model scores 0.7, about 70 in every 100 would be truly RTS.
Focal loss breaks this correspondence, because down-weighting the easy examples pushes every score towards the middle of the range and leaves the model under-confident \citep{mukhoti2020calibrating}.
This matters here because the map publishes the probability field itself and sorts each detection into a confidence band by its probability, so a high-score prediction must have a measurably higher hit rate.
We corrected it by temperature scaling \citep{guo2017calibration}, dividing the logit by a single parameter $T$ fitted on the validation set by minimising the binary cross-entropy against the labels.
A value of $T$ below one spreads scores away from 0.5, which is what an under-confident model needs, and the fitted $T = 0.512$ is itself the evidence that it was one.
The three seed probabilities $p$ are averaged and the temperature $T$ is applied to the logit of that mean,
\begin{equation}
\bar{p} = \tfrac{1}{3}\textstyle\sum_{k} p_k,
\qquad
p_{\mathrm{cal}} = \operatorname{sigmoid}\bigl(\mathrm{logit}(\bar{p})\,/\,T\bigr).
\label{eq:calibration}
\end{equation}
$T$ is fitted to the averaged probability rather than to each seed, since averaging and temperature scaling do not commute and the average is the quantity the deployed thresholds are applied to.
The selection threshold is then fitted on the same set and maximises object precision.
Section~\ref{sec:calibration} reports the fitted values and the evidence behind them, and Supplementary Table~S10 lists the deployment settings as applied.

Two design choices keep the output tractable at the scale of the 41.57 million tiles in the inference domain.
First, NDVI is not stored.
It is computed at read time from the red and near-infrared bands of the bulk Sentinel-2 composite, which avoids staging a separate auxiliary tile for every input tile, roughly 45\,TB.
Second, probabilities are stored as scaled 8-bit integers rather than 32-bit floats.
Each probability is multiplied by 250 and rounded, so the product can later be re-thresholded at any value in steps of 0.004, and each cloud-optimised GeoTIFF (COG) tile occupies about 8\,kB instead of roughly 570\,kB.

Inference ran on a Google Cloud virtual machine with eight NVIDIA A100 GPUs and one independent worker process per GPU.
The workers coordinate through cloud storage rather than a central scheduler, so the queue is self-balancing and resumes after interruption without a controlling process.
Supplementary Section~S5 details the shard, claim, heartbeat, and watchdog mechanics.

\subsection{Detection map and quality control}
\label{sec:postprocessing}
Two probabilities turn the field into a map.
The delineation threshold draws the boundary of each detection, and the selection threshold decides which detections the map keeps.
Detections are vectorised from the probability rasters at a delineation threshold of 0.30 with no fixed minimum size beyond a two-pixel technical minimum, and polygons split across tile seams are dissolved.
A residual-error diagnostic (Section~\ref{sec:testperf}) shows that reference objects peaking below 0.30 are not separable from background, so a lower threshold would admit false detections without recovering true ones.
Polygon boundaries are simplified by the Douglas-Peucker method \citep{douglas1973algorithms} at a tolerance of about one pixel.
Area and perimeter are measured geodesically on the WGS84 ellipsoid rather than in the projected raster, because the Web Mercator projection inflates area by about thirteen times at 74\,\textdegree{}N.
Each polygon carries its geodesic area, perimeter, centroid, mean and maximum probability, and a confidence band assigned from its maximum calibrated probability, with the selection threshold of 0.65 bounding the high band, 0.45 to 0.65 the medium, and 0.30 to 0.45 the low.

We also report an expected RTS area,
\begin{equation}
A_{\exp} = \sum_i p_i \, r^2 \cos^2\!\phi_i,
\label{eq:expected-area}
\end{equation}
summed over all valid pixels $i$, with $r = 4.77$\,m the projected resolution, $\phi_i$ the pixel latitude, and $\cos^2\!\phi$ the Web Mercator area correction.

Because no reference polygons exist for the 2025 imagery, we characterised the precision directly through a stratified expert review rather than assuming it from the validation record.
A sample of 280 polygons, stratified by confidence band and size, was rated by the first author as correct, incorrect, or uncertain.
The ratings were scored into a precision grid with Wilson confidence intervals, and the uncertain cases were set aside.
Because the sample is stratified rather than proportional to the map, it estimates precision within each confidence-by-size cell, not one precision for the whole map.
It also tested whether each confidence band needed a minimum size, by asking whether precision fell away in the smaller size classes.
Section~\ref{sec:product} reports the sample and its outcome.

\section{Results}
\label{sec:results}

The 172 controlled runs locked a recipe that reads only the PlanetScope visible bands and NDVI on an EfficientNet-B5 encoder.
NDVI was the one auxiliary channel adopted out of the twelve tested, and no combination of channels, no heavier fusion design, and no larger encoder beat it.
Of the three factors compared, the quantity of labelled data moved performance most, while model capacity saturated at EfficientNet-B5 and the advantage of a foundation-model encoder disappeared once the full label set was available.
Neither multi-scale training nor self-supervised Arctic pretraining met the criterion, and neither was adopted.
The final three-seed ensemble was tested once on held-out ecoregions.
It reached an object precision of 0.854 and an object F1 of 0.644 for the delivered map.
Deploying it over the 41.6 million tiles of the inference domain returned 19,068 high-confidence detections across 20.68 million km\textsuperscript{2}.
The subsections below report each of these in turn.

\subsection{Recipe build-up and channel ablation}
\label{sec:ablations}
\begin{figure}[!htbp]
\centering
\includegraphics[width=\linewidth]{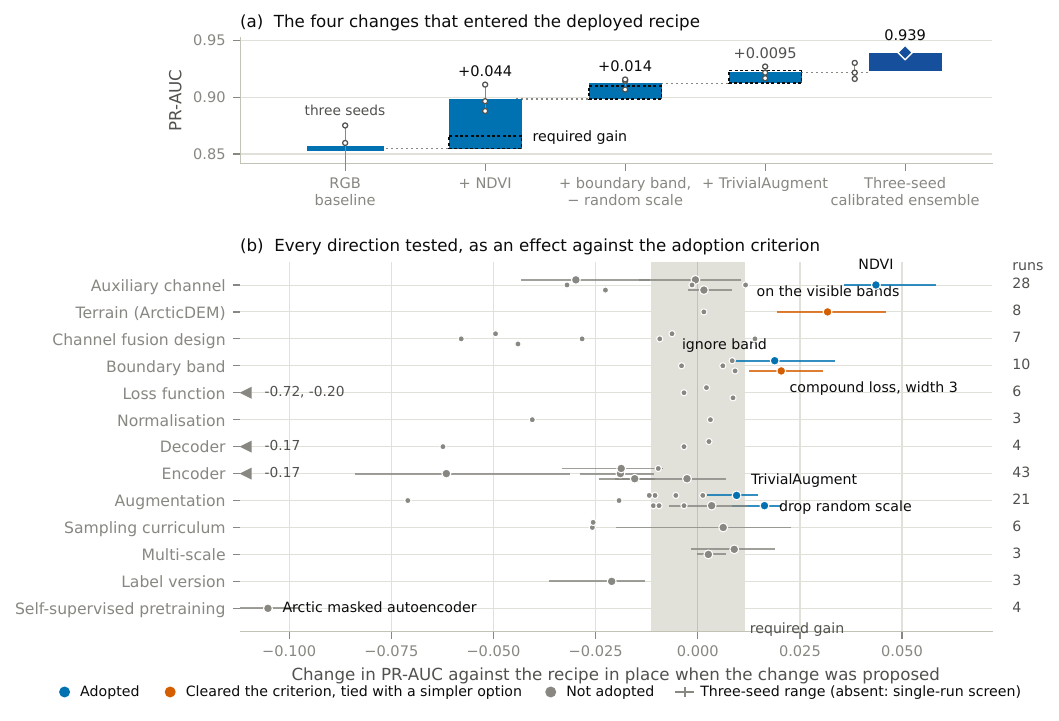}
\caption{Recipe build-up experiments. (a) The
changes that entered the deployed recipe, each drawn as the increment it added to the three-seed mean PR-AUC (geometric mean at prevalences 1:5, 1:10 and 1:20).
The boundary band and the removal of the random-scale augmentation are drawn as one step, because on the final split they were measured together.
Open circles are the three seed scores.
The dashed outline is the required gain of 0.0112, so a change met the criterion wherever the filled bar rises above it.
The final step has no spread because the deployed model averages the three seeds.
(b) Every direction tested, as a change in PR-AUC against the recipe in place when it was proposed.
The band spans the required gain on either side of zero, and a range bar marks a seed-confirmed result and its absence a single-run screen.
Each dot is a tested change rather than a single run, and the counts on the right are the runs behind each direction including controls.
}
\label{fig:ablation}
\end{figure}

The recipe was built up one change at a time, each held to the adoption criterion of Equation~\ref{eq:gate}.
Figure~\ref{fig:ablation}a shows the progression, and Figure~\ref{fig:ablation}b places every direction tested on a single axis against that criterion.
Adding NDVI to the visible bands, setting the boundary band to ignore, removing the random-scale augmentation, and replacing the hand-tuned colour augmentation with TrivialAugment together fixed the convolutional recipe at 0.922, and Supplementary Section~S1 and the run table \texttt{supp\_runs.csv} record the full experiment programme.
Three of these four met the adoption criterion.
TrivialAugment fell short of it by 0.0017, at a three-seed mean gain of 0.0095 against the required 0.0112, though it improved in all three seeds.
Because the criterion could not separate it from the hand-tuned colour stack it replaced, the tie-break of Section~\ref{sec:experiments} applied, and the parameter-free policy was kept as the less tuned of the two.
No other change among the 172 runs entered the deployed recipe.
The compound loss shows the same rule from the other side.
Both it and the adopted focal loss met the criterion against the recipe that had no boundary band, gaining 0.020 and 0.019 respectively, so the improvement came from the boundary band rather than from the loss.
The two then differed from each other by 0.0016, a seventh of the required gain, at 0.812 for the compound loss against 0.810 for the focal loss on three seeds, and the focal loss was kept as the simpler of the two.
Every other group confirmed a simpler or cheaper default and rejected the alternative tested against it.

Every auxiliary channel helped on its own, but adding more than one did not help and often hurt.
Candidate channels were first screened one at a time in single runs against a single-run RGB control of 0.830, and only the best of them was then repeated on three seeds for the adoption decision (Supplementary Section~S1).
Added alone to the RGB bands in these single runs, the NBR, the embedding prototype, the tasselled-cap pair, and the SE embedding components raised the score to between 0.847 and 0.874, and NDVI raised it the most, to 0.888.
Only NDVI was then seeded, and on three seeds it raised the score from 0.855 to 0.898, a gain of 0.044 that is nearly four times the required gain.
These gains did not combine.
Greedy addition of a second group to NDVI gave nothing, with the tasselled-cap, embedding-component, and prototype pairs all within noise of NDVI alone at 0.898 to 0.900, while adding the NBR degraded the score to 0.856, close to that channel on its own.
The full eight-band auxiliary stack of Section~\ref{sec:channels} likewise scored 0.869 on three seeds, below NDVI alone.
The four heavier fusion designs of Section~\ref{sec:experiments} also lost to the single channel.
Zero-initialising the auxiliary stem of the eight-band stack scored 0.890, within noise of the plain stack, the input channel-attention gate collapsed to 0.827, the dual encoder with late fusion reached 0.818, and residual cross-modal attention reached 0.848, all below NDVI.
Section~\ref{sec:disc-representation} interprets this pattern.

The ArcticDEM terrain group followed the same pattern.
Its scores are not directly comparable with those above because the runs were restricted to the DEM-covered tiles (Section~\ref{sec:channels}).
Added to the visible bands alone, terrain raised the three-seed mean score by 0.0318 with the same sign in all three seeds, 2.8 times the required gain, and was the only auxiliary channel other than NDVI to meet the criterion.
Added on top of NDVI, it raised the score by 0.0015, an eighth of the required gain, against 0.0498 for NDVI added to the visible bands on the same tiles.
One caveat bounds this: the comparison on top of NDVI rests on a single seed, while the seed spread of the terrain runs was 0.0178, so the experiment fails to demonstrate a benefit rather than showing that none exists.
Additionally, the ArcticDEM covers only 71.8\% of the inference domain by area, so adopting terrain would have cost more than a quarter of the map for a gain the experiment could not demonstrate.
Accordingly, terrain is not part of the deployed model (Supplementary Section~S3).

\subsection{Labels, capacity and pretraining}
\label{sec:levers}
\begin{figure}[!htbp]
\centering
\IfFileExists{figures/fig07_capacity.pdf}{%
  \includegraphics[width=\linewidth]{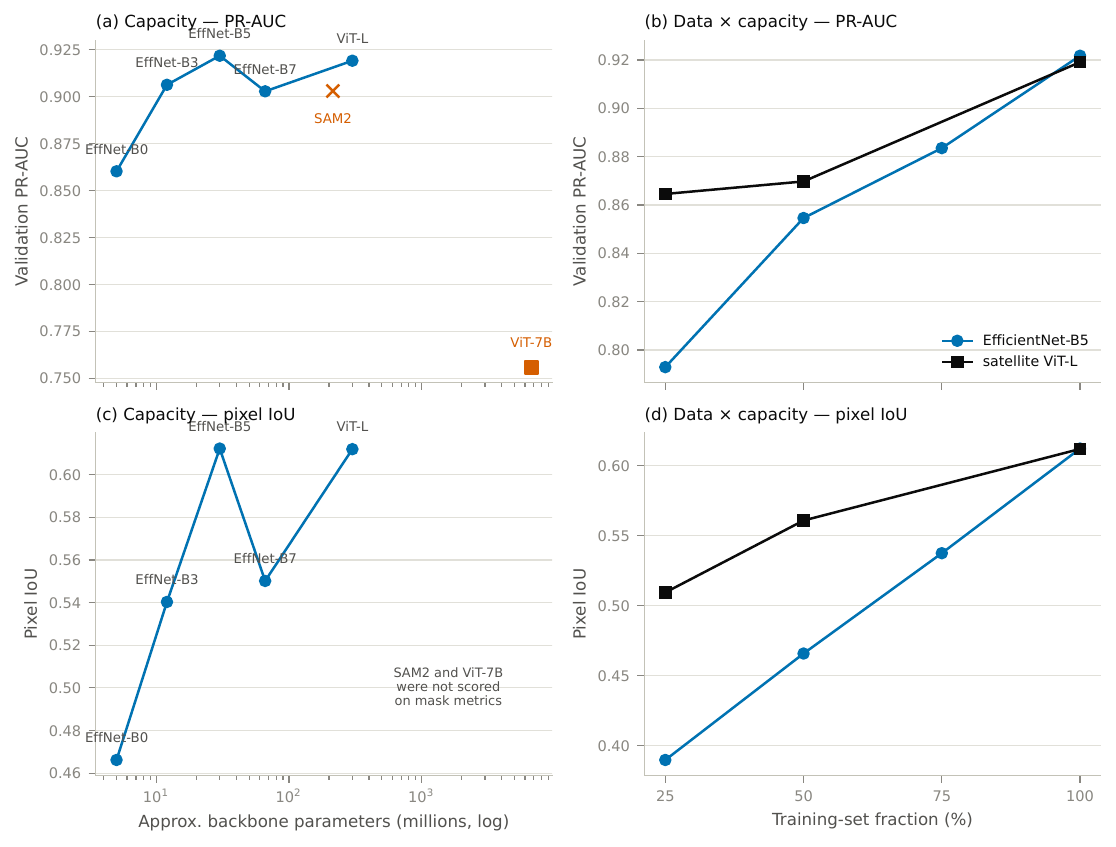}}{%
  \fbox{\parbox{0.9\linewidth}{\centering\vspace{3cm}\textbf{[run
  figures/src/fig07\_capacity.py to generate fig07\_capacity.pdf]}\vspace{3cm}}}}
\caption{Impact of model capacity and training data quantity on model performance.
(a) Performance against backbone size.
(b) Performance against the quantity of training data.
(c) and (d) repeat (a) and (b) with pixel IoU in place of PR-AUC.
Each column shares one horizontal axis, so a given encoder or label budget sits at the same position in both of its panels.
Backbone parameter counts are nominal published architecture sizes rather than measured ones.}
\label{fig:capacity}
\end{figure}

Three factors govern this task: the quantity of labelled data, the capacity of the model, and the pretraining behind the encoder (Figure~\ref{fig:capacity}).

Data volume matters most of the three, and it is the only factor that raises the performance ceiling rather than approaching it.
A data-scaling study at the final recipe raised the score from 0.793 at a quarter of the training tiles to 0.922 at the full set, a gain of 0.129 in PR-AUC and 0.222 in pixel IoU, more than ten times the required gain (Figure~\ref{fig:capacity}b).
The training-to-validation pixel IoU gap narrowed from about 0.32 to 0.08 as data were added, and the curve was still rising at the full set, so the task remains data-limited rather than saturated.

Model capacity, by contrast, does matter, but only up to a modest size.
Within the EfficientNet family at the fixed RGB and NDVI input, the score rose from 0.860 at EfficientNet-B0 to 0.922 at EfficientNet-B5, a gain of 0.062 in PR-AUC and 0.146 in pixel IoU, and then saturated: the larger EfficientNet-B7 and the satellite-pretrained DINOv3 ViT-L, an order of magnitude larger than EfficientNet-B5, held between 0.90 and 0.92 (Table~\ref{tab:encoders}, Figure~\ref{fig:capacity}a).
The foundation model tied EfficientNet-B5 (0.919 vs 0.922) on every metric while costing about four times as much to run, so the convolutional network was kept.
Among decoders, UNet++ was at least as good as the alternatives.
SAM2 reached a three-seed mean of 0.903, 0.019 short of EfficientNet-B5 on an encoder about four times as costly.
The 6.7-billion-parameter DINOv3 reached 0.756, short of EfficientNet-B5 at about 3.3 times the training cost.
No encoder larger than EfficientNet-B5 exceeded it on any metric (Figure~\ref{fig:capacity}a).

The difference in performance across encoders depended on the quantity of training data.
Repeating the data-scaling study with the foundation model in place of the convolutional network makes the size of the encoder advantage a function of the label budget (Figure~\ref{fig:capacity}b).
At a quarter of the labels, the foundation model scored 0.865 against the convolutional network's 0.793, an advantage of 0.072 in PR-AUC.
At half the labels the advantage fell to 0.015, and at the full set it vanished, 0.919 against 0.922.
The NDVI spectral prior follows the same pattern, adding 0.044 to the convolutional network but only 0.001 to the foundation model.
Section~\ref{sec:disc-representation} interprets the three factors together.

Pixel IoU and pixel F1 move with PR-AUC for all three factors, so the findings above hold across metrics with one exception (Supplementary Table~S1).
The object-level F1 decreased from 0.521 to 0.491 when NDVI was added.
The RGB and larger models were more precise at the object level, because a higher-recall probability field recovers more RTS at the cost of more false detections.
This is a precision-recall trade-off at the object level.
Adoption was decided on the threshold-free selection metric, and the object-level balance between precision and recall is then fixed by the detection settings rather than by the channel, so a channel that raises recall does not commit the deployed map to it.
Section~\ref{sec:testperf} reports the object metrics at fixed detection settings.

\begin{table}[!htbp]
\centering
\caption{Encoders compared at fixed input representation (RGB and NDVI,
final split), under the final recipe with identical validation labels.
Within the EfficientNet family, capacity helps up to B5 and then saturates, and the foundation-model encoders, far larger, do not beat it at the full label budget.}
\label{tab:encoders}
\footnotesize
\setlength{\tabcolsep}{3pt}
\begin{tabular}{@{}l l r c c@{}}
\toprule
Encoder & Capacity & PR-AUC & IoU / obj.\ F1 & Cost \\
\midrule
\multicolumn{5}{@{}l}{\emph{EfficientNet family}} \\
EfficientNet-B0 & very small & 0.860 & 0.466 / 0.372 & $<$1$\times$ \\
EfficientNet-B3 & small & 0.906 & 0.540 / 0.384 & $<$1$\times$ \\
EfficientNet-B5 (deployed) & medium & \textbf{0.922} & \textbf{0.612} / \textbf{0.438} & 1$\times$ \\
EfficientNet-B7 & large & 0.903 & 0.550 / 0.403 & $\sim$2$\times$ \\
\addlinespace[3pt]
\multicolumn{5}{@{}l}{\emph{Foundation-model encoders}} \\
DINOv3 ViT-B (web) & medium & 0.912 & --- & --- \\
DINOv3 ViT-L (satellite) & large & 0.919 & \textbf{0.612} / \textbf{0.438} & $\sim$4$\times$ \\
DINOv3 ViT-L (satellite) + Arctic MAE & large & 0.814 & --- & $\sim$4$\times$ \\
DINOv3 ViT-7B (frozen encoder) & very large & 0.756 & 0.343 / 0.382 & $\sim$3.3$\times$ \\
SAM2 (Hiera) & large & 0.903 & --- & $\sim$4$\times$ \\
\bottomrule
\end{tabular}
\tabnote{PR-AUC is the three-seed geometric mean at prevalences 1:5, 1:10 and 1:20 on the validation set, while pixel IoU and object F1 are three-seed means at the fixed reporting threshold.
Relative cost is wall-clock time against the deployed EfficientNet-B5, at inference for the ViT encoders and at training for the frozen 7B encoder.
The Arctic MAE row is the satellite DINOv3 ViT-L after continued masked-autoencoder pretraining (Section~\ref{sec:experiments}), which lowers the score to 0.814.}
\end{table}

\subsection{Multi-scale and self-supervised training}
\label{sec:pretraining}
Neither multi-scale training nor self-supervised pretraining was included in the final model recipe.

The multi-scale experiments of Section~\ref{sec:experiments} passed their first two criteria but failed the third.
Training with the half-resolution copy held the native-resolution score at 0.924 against the 0.922 baseline, and it gave the model a usable half-resolution capability that the baseline lacks, since a model trained only at native resolution collapses when applied at half resolution.
Fusing native and half-resolution predictions at inference, however, traded recall for precision and never raised object recall above the native scale alone.
The deployed system therefore runs at a single scale, while the trained half-resolution capability is retained for coarser sensors.

The self-supervised pretraining (Section~\ref{sec:experiments}) failed all criteria.
Pretraining itself converged, with the reconstruction loss falling from 1.016 to 0.076 over the 80 epochs.
Under the final recipe, the unadapted satellite DINOv3 ViT-L then reached 0.919 against the EfficientNet-B5 baseline at 0.922.
After masked-autoencoder pretraining on the 295,429 unlabelled Arctic tiles, the same encoder under the same recipe fell to a three-seed mean of 0.814 (Table~\ref{tab:encoders}).
Section~\ref{sec:disc-representation} interprets this result.

\subsection{Temperature scaling and detection settings}
\label{sec:calibration}
\begin{figure}[!htbp]
\centering
\includegraphics[width=\linewidth]{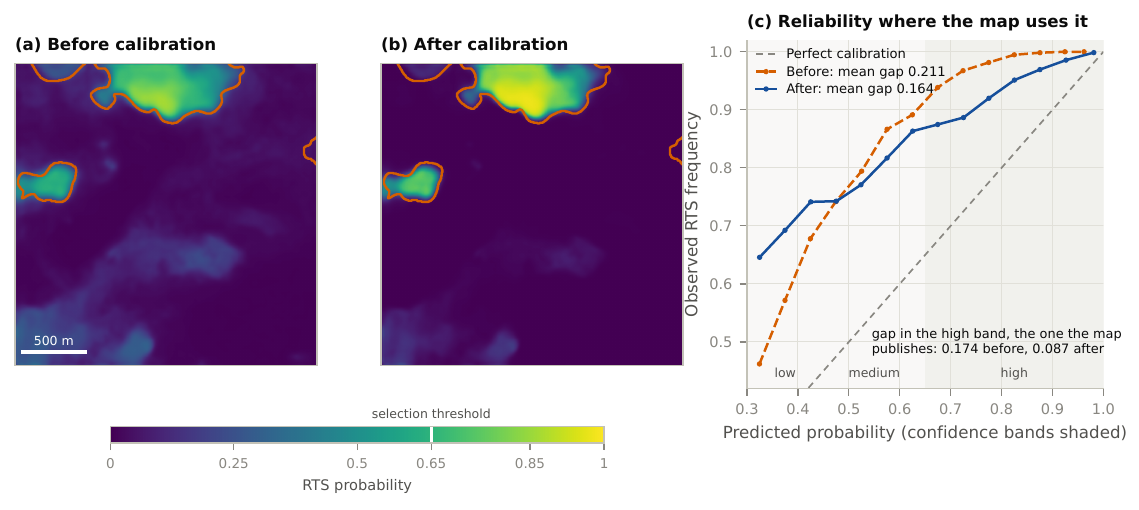}
\caption{The effect of temperature scaling on the probability field.
(a, b) One held-out test tile, 2.4\,km on a side, as the raw three-seed
ensemble mean and after temperature scaling, on one shared scale with the
reference RTS outline in orange and the selection threshold of 0.65 marked on
the colour bar. Scaling raises the probability inside the RTS and lowers it
across the surrounding tundra.
(c) Calibration gap over the held-out test pixels, before and after, with the
confidence bands of the delivered map shaded.
The temperature was fitted on the validation regions, while the gap is measured
on the test regions.}
\label{fig:probmap}
\end{figure}

Averaging the three seed models at the probability level raised the validation score to 0.939, which is 0.009 above the best of the three seeds and 0.017 above their mean, and, more importantly, removed the seed-to-seed variance.
Temperature scaling (Equation~\ref{eq:calibration}) fitted a value of 0.512, below one, which confirms that the uncalibrated ensemble was under-confident.
It halved the calibration gap in the high-confidence band, from 0.174 to 0.087 (Figure~\ref{fig:probmap}).
That gap measures how far the probability the map states sits from the frequency with which those pixels are in fact RTS, so a stated probability in the high band now falls within 0.087 of the rate a user would observe.
The selection threshold that maximised object precision on the validation set was 0.65.
At this threshold, the object precision was 0.611 and the recall was 0.439.
An alternative would be the selection threshold that maximised object F1, 0.30.
At this threshold, the F1 was 0.567 at a precision of 0.489 and a recall of 0.674.
Choosing 0.65 therefore gave up 0.235 of object recall and 0.056 of object F1 to gain 0.122 of object precision, a trade we accepted because a false detection costs this map more than a missed one (Section~\ref{sec:disc-limitations}).

\subsection{Held-out test performance}
\label{sec:testperf}
\begin{figure}[!htbp]
\centering
\includegraphics[width=\linewidth]{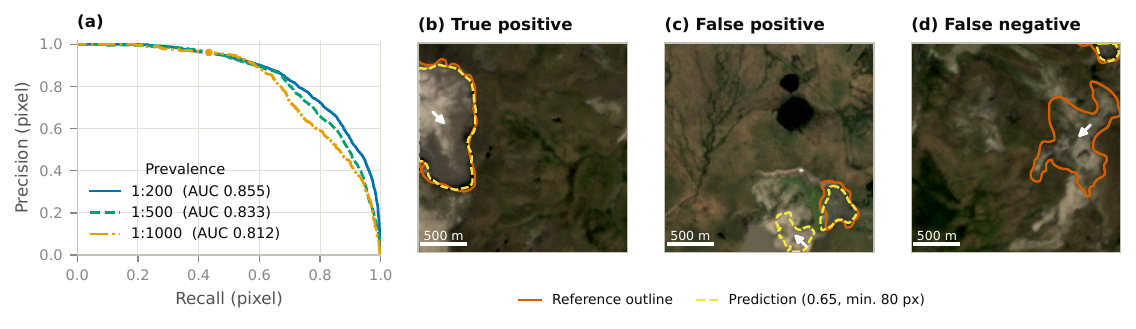}
\caption{Held-out test performance of the final three-seed ensemble.
(a) Precision-recall curves over the pooled test pixels at prevalence ratios of 1:200, 1:500, and 1:1000 (PR-AUC 0.855, 0.833, and 0.812); markers show the selection threshold of 0.65.
(b--d) Example detections at the settings used for comparison: a true positive, a false positive, and a false negative, with reference outlines in orange and predictions dashed in yellow.
Each panel is one whole 512 by 512 pixel test tile, and the arrow marks the RTS the tile is meant to highlight.
}
\label{fig:test}
\end{figure}

The final ensemble was scored once on the held-out test regions.
Table~\ref{tab:testmetrics} reports the test metrics, and Figure~\ref{fig:test} shows the precision-recall curves with example detections.

\begin{table}[!htbp]
\centering
\caption{Held-out test metrics, scored on one stored set of predictions.}
\label{tab:testmetrics}
\small
\begin{tabular}{lc}
\toprule
Metric & Value \\
\midrule
Object precision / recall / F1, delivered map & 0.854 / 0.516 / 0.644 \\
Object precision / recall / F1, frozen point & 0.584 / 0.437 / 0.500 \\
Pixel IoU / F1 & 0.432 / 0.604 \\
PR-AUC (1:200 / 1:500 / 1:1000) & 0.855 / 0.833 / 0.812 \\
\bottomrule
\end{tabular}
\tabnote{Both object rows come from the same stored predictions, with matched
pairs requiring an intersection over union of at least 0.3. The delivered-map
row applies the rule of Section~\ref{sec:postprocessing}, which outlines
detections at a probability of 0.30 and selects them by peak probability with
no size filter.}
\end{table}

The test PR-AUC was 0.855, 0.833, and 0.812 at prevalence ratios of 1:200, 1:500, and 1:1000 (Figure~\ref{fig:test}a).
These values are lower than the validation ensemble score of 0.939.
Still, the two are not directly comparable because the validation score was measured at the ratios of 1:5, 1:10, and 1:20 used for model selection.
The object metrics matched the validation set closely, with object F1 moving from 0.511 to 0.500 and object precision from 0.611 to 0.584, which shows the detection settings generalised rather than overfitting the validation data.

A residual-error diagnostic asks whether any threshold could select an object at all.
Of the 132 validation reference objects, 80 are detected at the selection threshold of 0.65 and a further 15 carry recoverable signal between 0.30 and 0.65.
The remaining 37, more than a quarter, fall into the undetectable fraction of Section~\ref{sec:evaluation}.
On the held-out test set that fraction is 0.223, so no threshold could lift recall above about 0.78.
Section~\ref{sec:disc-representation} interprets this limit.

\subsection{The circumpolar detection map}
\label{sec:product}
Deploying the final three-seed ensemble over the 41.57 million tiles of the inference domain produces a circumpolar map of RTS detections.
Each GPU sustained about 27 tiles per second, and the run completed in about 2.3 days.
It was coordinated through cloud storage, so it resumed without loss across preemptions (Section~\ref{sec:inference}).
The run cost about 440 A100-hours, against at least 2,100 A100-hours for the whole experiment programme that preceded it, so mapping the Arctic once was roughly a fifth of the cost of deciding how to map it (Supplementary Table~S10).
Its primary output is the calibrated probability field, from which detections are vectorised at any threshold.
At the delineation threshold of 0.30, the field yields a candidate pool of 60,167 polygons covering 688.2\,km\textsuperscript{2}.
Ranking these by their peak calibrated probability places 19,068 in the high-confidence band, at or above the selection threshold of 0.65, covering 529.7\,km\textsuperscript{2}, with 11,865 medium and 29,234 low-confidence polygons below it.
Prioritising precision (Section~\ref{sec:calibration}) thus sets aside roughly two-thirds of the candidate pool, which the probability field retains for any user who prefers a more permissive cut.
These are model detections rather than a census-verified inventory, so their reliability is characterised here from a stratified sample.
The probability field and both bands are browsable as provisional model detections through a public Google Earth Engine application at \url{https://abruptthawmapping.projects.earthengine.app/view/south-rts-map}.

The stratified expert review of Section~\ref{sec:postprocessing} rated 280 polygons spanning the confidence bands and size classes, returning 65 correct, 152 incorrect, and 63 uncertain, so 217 verdicts were scored.
Those counts are not a map-level precision, because the sample deliberately over-represents the low-confidence and small-object cells where errors were expected to concentrate.
Read per cell, the review gives a sampled precision for the high-confidence detections ranging from 0.54 to 0.90 across size classes.
It is highest, at 0.90, in the smallest measured class of 500 to 2,000\,m\textsuperscript{2}, so precision does not fall with size.
We therefore applied no size filter to the final map.

\subsection{Distribution of the detected RTS}
\label{sec:distribution}
\begin{figure}[!htbp]
\centering
\includegraphics[width=\linewidth]{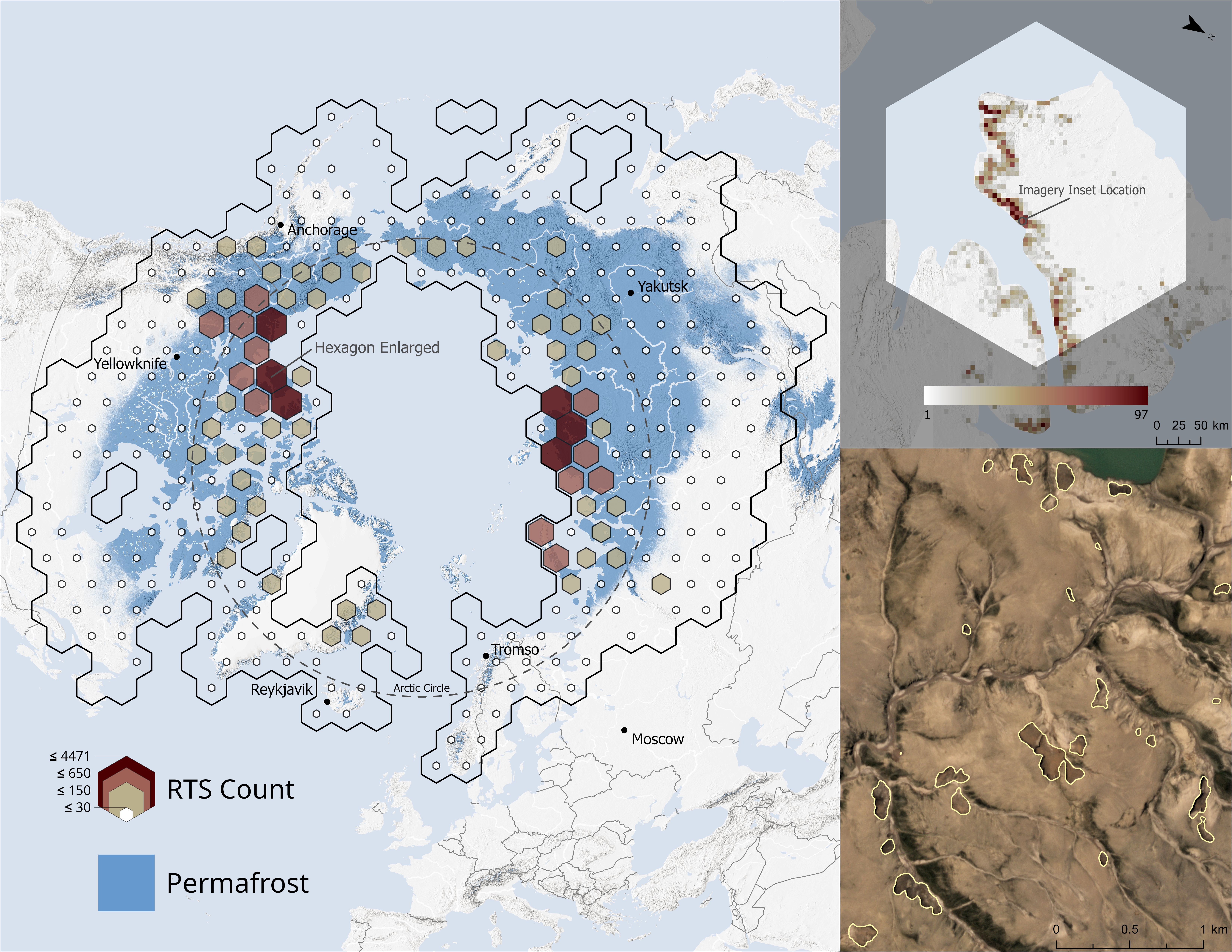}
\caption{Pan-Arctic map of high-confidence retrogressive thaw slump
detections. The hexagonal grid covers the 20.68 million km\textsuperscript{2} inference domain (black), and each cell is shaded and scaled by the number of RTS it holds.
Each hexagonal cell is 100,000 km\textsuperscript{2}.
Top right: a zoomed view of one hexagonal cell over part of Banks Island, with colour indicating RTS count over 5 by 5 km pixels.
Bottom right: a zoomed view of one RTS hotspot on Banks Island, showing PlanetScope imagery with detections outlined in yellow.
}
\label{fig:product}
\end{figure}

The detections are concentrated in the high Arctic (Figure~\ref{fig:product}).
Nine in ten lie north of 65\,\textdegree{}N and seven in ten north of 70\,\textdegree{}N, at a median latitude of 72.1\,\textdegree{}N, while only 2.6\% lie south of 60\,\textdegree{}N.
They are divided between the two land masses, with 56\% in North America and 44\% in Eurasia.
Individual detections are small, with a median area of 1.6\,ha and a 90th percentile of 6.1\,ha (Table~\ref{tab:product}).
Density varies by more than an order of magnitude across the 53 ecoregions in which detections occur.
The Canadian Middle Arctic tundra and the Taimyr-Central Siberian tundra alone hold 60\% of all detections, at 6.1 and 5.8 detections per thousand km\textsuperscript{2}.
The next highest densities are 3.8 in the Ogilvie-MacKenzie alpine tundra and 2.6 in the Russian Arctic desert.
Repeat surveys have already reported rapid RTS growth in these same regions \citep{nitze2025darts, dai2025volumetric, lewkowicz2019extremes}, so the map reproduces a known pattern and places it on one consistently searched domain.
Section~\ref{sec:disc-comparison} sets it beside six published RTS inventories.

\begin{table}[!htbp]
\centering
\caption{Pan-Arctic detection statistics from the 2025 inference run. These are
raw, high-confidence model detections with a sampled precision, not a
census-verified inventory.}
\label{tab:product}
\small
\begin{tabular}{@{}l l@{}}
\toprule
Statistic & Value \\
\midrule
Candidate polygons (threshold 0.30) & 60{,}167 \\
Candidate area (km\textsuperscript{2}) & 688.2 \\
High-confidence detections (peak prob.\ $\geq 0.65$) & 19{,}068 \\
High-confidence area (km\textsuperscript{2}) & 529.7 \\
Median / 90th percentile high-confidence area (ha) & 1.6 / 6.1 \\
Median peak detection probability & 0.88 \\
Sampled precision, high-confidence (by size class) & 0.54--0.90 \\
Expected RTS area, threshold-free (km\textsuperscript{2}) & 1{,}037.4 \\
Ecoregions with high-confidence detections & 53 \\
\bottomrule
\end{tabular}
\tabnote{Areas are geodesic measures on the WGS84 ellipsoid
(Section~\ref{sec:postprocessing}), computed from the delivered candidate
inventory.
Counts, areas, and the size, probability and ecoregion statistics beyond the
headline totals describe the high-confidence band. Ecoregions are counted
against the full set intersecting the inference domain (Section~\ref{sec:domain}).}
\end{table}

\section{Discussion}
\label{sec:discussion}
The present study contributes in three ways.
First, it searches the circumpolar region covering 20.68 million km\textsuperscript{2}.
This is more than a hundred times the largest region mapped by the change-based methods \citep{huang2021beiluhe, ma2025imagestacks, maier2025elevation}.
It also releases the ground it searched, so that an absence of detections can be told apart from an absence of search.
For inference and map production at this scale, it reports the end-to-end execution and optimisation details for reproducibility and quality assessment.
Second, it compares ten groups of RTS segmentation model factors under one controlled protocol through 172 experiments (Supplementary Section~S1).
The groups range from loss and model size to pretraining, data fusion and input channels.
To our knowledge, no earlier RTS study has made this comprehensive comparison and scored at the true RTS prevalence.
Third, it reports lessons learned for mapping objects under extreme class imbalance with scarce labels.
Its mitigations for both could extend to other Earth Observation tasks.

\subsection{Comparison with published inventories}
\label{sec:disc-comparison}
\begin{figure}[!htbp]
\centering
\includegraphics[width=0.83\linewidth]{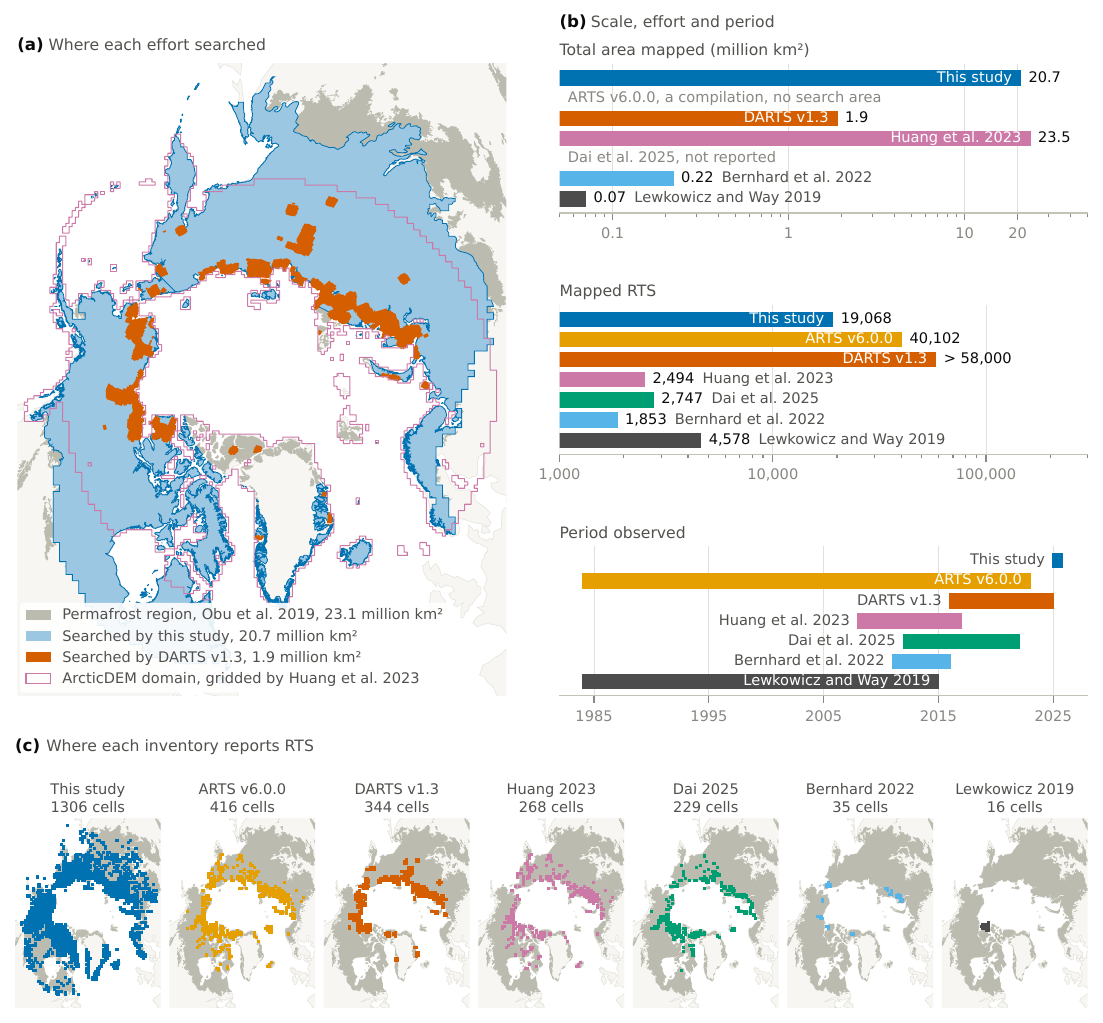}
\caption{This study beside six published retrogressive thaw slump inventories.
(a) Where each effort searched, over the permafrost region of \citet{obu2019northern} in grey. The two fills are the only efforts that release the ground they searched.
(b) Scale, effort and period. Area mapped is measured from the released coverage layer where one exists and quoted from the paper otherwise, so the longest bar counts grid cells over ocean and ice sheet as well as land.
(c) Where each inventory reports RTS, on one 100 km lattice. A cell counts once an inventory reports a single RTS inside it.
Sources are \citet{nitze2025darts}, \citet{yang2025arts}, \citet{huang2023identifying}, \citet{dai2025volumetric}, \citet{bernhard2022assessing} and \citet{lewkowicz2019extremes}.}
\label{fig:inventorycomp}
\end{figure}

This study releases the largest continuous search of the circumpolar permafrost region so far, and one of only two efforts that publish the ground they searched.
It extends the inventories that came before it rather than replacing them.
Figure~\ref{fig:inventorycomp}a sets the 20.7 million km\textsuperscript{2} searched here against the 1.9 million km\textsuperscript{2} of coverage released by DARTS v1.3, the other such effort \citep{nitze2025darts}.
Read against the permafrost zonation of \citet{obu2019northern} rather than against our domain (Section~\ref{sec:domain}), the two together cover 85\% of the region drawn in the figure and leave 15\% unsearched, and the DARTS footprint falls almost entirely inside the ground searched here.
\citet{huang2023identifying} report a larger total of 23.5 million km\textsuperscript{2}, but that number counts 20 km grid cells over the whole ArcticDEM domain, ocean and ice sheet included, rather than land inside the permafrost region.
That domain is outlined in Figure~\ref{fig:inventorycomp}a, and the tiles behind it enclose 22.2 million km\textsuperscript{2}, so the grid counts rather more ground than the domain itself holds.

The mapped RTS counts are not directly comparable (Figure~\ref{fig:inventorycomp}b).
Three differences explain why.
This map covers one 2025 growing season, whereas DARTS accumulates features over 2016 to 2025 and ARTS over 1984 to 2023.
DARTS also counts active-layer detachment slides alongside RTS, so its total is not an RTS count.
Finally, the 19,068 detections are the high-confidence band of a 60,167-polygon candidate pool.
That is what choosing precision over recall costs (Section~\ref{sec:disc-limitations}).
What the map adds instead is spatial reach.
Figure~\ref{fig:inventorycomp}c counts each cell of a 100 km lattice once an inventory reports a single RTS inside it.
On this lattice, this study reports detections in 1,306 cells against 416 for ARTS v6.0.0 and 344 for DARTS v1.3, and no other inventory reaches 300.
Across all seven efforts, 1,452 cells hold a reported RTS, and only 49 of them fall outside the domain searched here.

Our object F1 appears lower than most scores in the RTS-mapping literature.
However, the numbers are not directly comparable because of what those scores are measured on.
Three axes govern whether they are comparable at all: the geography of the test set, the prevalence of the evaluation set, and the rule by which a detection counts as correct.
Supplementary Table~S3 places representative studies along them.
On prevalence the literature is consistent.
The highest reported scores are measured on RTS-bearing imagery rather than at true spatial occurrence across Arctic landscapes.
\citet{nitze2021testing} report their maximum IoU of 0.39 to 0.58 for models trained and validated on tiles containing RTS, thereby undersampling the background class.
\citet{huang2020using} evaluate over their full study area, but every reference polygon was also used in training.
\citet{yang2023mapping} score 69 RTS-bearing test images alongside a curated negative set, and \citet{zeng2026multisource} score 179 chips holding 298 reference objects.
The matching rules also differ widely.
A detection counts as correct at an IoU of 0.5 in \citet{huang2020using}, at any spatial overlap in \citet{zeng2026multisource}, and at an IoU of 0.3 here.

When \citet{yang2023mapping} added a curated negative set to their test data, object F1 fell from 0.81 to 0.66, with 46 of 61 false positives arising from the negatives, and when \citet{nitze2021testing} trained on the full negative pool, precision rose while recall fell.
We interpret this as the same effect that the prevalence scaling captures.
A score measured where RTS are common will likely not transfer to a landscape where they occupy under one per cent of tiles, because the false positives that dominate an operational map are drawn from the negatives that such an evaluation excludes.

Only \citet{nitze2025darts} evaluates its map the way we do, against independent reference data over whole regions rather than over RTS-positive tiles.
The circumpolar DARTS product reports an internal held-out F1 near 0.80, but when validated against independent regional inventories its agreement falls to an F1 between 0 and 0.70, with precision consistently exceeding recall \citep{nitze2025darts}.
Our delivered map scores an object F1 of 0.644 at a precision of 0.854 on a held-out ecoregion, which lies in the upper part of that independent-validation band.
We reach this using only RGB and NDVI, whereas DARTS additionally ingests elevation and multi-temporal trend layers.
Elevation did not improve on NDVI under our own protocol (Supplementary Section~S3), and trend layers were not tested.
The comparison also frames our central finding.
Studies that add elevation, thermal and multi-temporal channels report fusion gains \citep{li2025multimodal, zeng2026multisource}.
Yet, none benchmark those gains against a strong single-channel baseline under a matched protocol that blocks by region and scores at the true rarity of RTS.
Our ablation supplies that control and finds that, in our case, the gains do not survive it (Section~\ref{sec:disc-representation}).

Other strategies published for mapping RTS look at change over time rather than at a single image \citep{huang2021beiluhe, ma2025imagestacks, maier2025elevation}.
These methods directly measure how fast RTS grow in a single run.
From elevation change, they could also measure the volume removed.
Our method, run on successive annual basemaps, would measure growth in area by comparing the detections it finds in each year.
Two of their results also relate to our findings.
NDVI carries the RTS signal through time in the image-stack method \citep{ma2025imagestacks}, as it does through space in our detector.
This is consistent with NDVI being the auxiliary channel our ablation adopted (Section~\ref{sec:ablations}).
Elevation change in the interferometric method \citep{maier2025elevation} also carries the RTS signal, whereas static terrain from ArcticDEM added nothing to detection once NDVI was present (Supplementary Section~S3).
This contrast is consistent with the information elevation holds about active RTS lying in its change rather than its shape.

\subsection{Channel redundancy and the label limit}
\label{sec:disc-representation}

Each auxiliary channel improved detection alone, yet no pair of them beat NDVI alone.
Pairs with NBR were worse than NDVI alone (Section~\ref{sec:ablations}).
We interpret this as noisy redundancy rather than interaction.
The candidate channels all respond to similar physical phenomena of vegetation loss and increased moisture on the barren floor and headwall that define an active RTS.
The screening correlations in Supplementary Section~S2 illustrate this well, with NBR and NDVI reaching a rank correlation of 0.75 near slump boundaries.
Once NDVI presents that signature directly to the network, a second channel provides little new information while increasing the number of input dimensions the network must fit from about 1,700 positive tiles.
More input channels at a fixed number of training examples can lower accuracy, an effect known since before deep learning as the Hughes phenomenon, or the curse of dimensionality \citep{hughes1968accuracy}.
The same has been reported for convolutional landform mapping in permafrost terrain, where models trained on all available bands lost accuracy against well-chosen subsets \citep{bhuiyan2020bands}.
A further, more speculative reading is that NBR, the channel whose combinations degraded, also carries the noisiest copy of the shared signal, since its shortwave-infrared band has a native resolution of 20\,m.
ArcticDEM behaved similarly, as it helped on its own but was redundant once NDVI was present (Section~\ref{sec:ablations}).
This could be mainly due to the temporal offset of the DEM, which adds considerable noise to the input.
The practical lesson is that channels for a rare-target task should be selected jointly rather than accumulated on an individual basis, and that gains reported for multi-source fusion frameworks \citep{li2025multimodal, zeng2026multisource} deserve to be tested against the strongest single channel under the same protocol.

The ceiling on this task is set by the labels, not by the model.
What matters is how many labels there are and how much of the variation they cover.
The residual-error diagnostic of Section~\ref{sec:testperf} provides the direct evidence that more than a quarter of the validation reference objects peak below a probability of 0.30.
These objects carry less detectable optical signal in the input, so no increase in encoder capacity can recover them.
The labels are not the source of this residual error, since the first author reviewed and edited every training and evaluation label for internal consistency, so the objects that carry no signal are limited by the imagery rather than by annotation noise.
This is why at the full label budget, a satellite foundation encoder an order of magnitude larger than EfficientNet-B5 only ties it, and why even a 6.7-billion-parameter encoder still reaches only 0.756 (Table~\ref{tab:encoders}).
Model capacity approaches this ceiling but does not raise it.
Only more labelled data does, and the data-scaling curve of Section~\ref{sec:levers} was still rising at the full set, suggesting it remains data-limited.
The advantage of a foundation model is significant when labels are scarce.
This is the same mechanism seen from the other side: pretraining supplies, from hundreds of millions of unlabelled images, the signal the small network would otherwise learn from labels, and that advantage is spent once the labels are available.

The self-supervised result further strengthens the argument above.
The encoder comparison varied in capacity and pretraining source but not in domain-adapted weights, so it remained possible that the foundation encoders fell short only because their pretraining imagery was foreign to the Arctic tundra.
Continuing their pretraining on 295,429 Arctic tiles removed exactly that objection and made the model worse, a fall of 0.105 with no seed improving (Section~\ref{sec:pretraining}).
Our leading interpretation is catastrophic forgetting.
The reconstruction objective rewards features that predict masked pixels, and optimising it appears to have overwritten the discriminative features the satellite checkpoint already carried (DINOv3 is pretrained on the SAT-493M dataset, Maxar RGB ortho-rectified imagery at 0.6\,m resolution), a reading consistent with the adapted models reaching their fine-tuning peaks later than the controls.
The scope of the finding is limited to a four-channel corpus from the inference domain and a single masking schedule on PlanetScope, so it does not rule out self-supervision for this task in general.
It does show that domain-adapted reconstructive pretraining is not a free improvement in this regime.
Nonetheless, satellite pretraining is a mitigation for scarce labels (Section~\ref{sec:levers}), and further reconstructive pretraining on a small, lower-resolution corpus overwrote that advantage rather than adding to it, leaving labelled data as the limit.

\subsection{Measuring accuracy for rare targets}
\label{sec:disc-rigour}
Two measurement practices matter more here than any modelling choice, and both transfer to other rare-object mapping.
The first is to split the data by region rather than by tile.
Spatial autocorrelation between adjacent training and test samples is a documented cause of inflated accuracy in both ecological mapping and convolutional segmentation \citep{roberts2017cross, ploton2020spatial, wang2023spatialplus, kattenborn2022spatially, meyer2022global}.
We confirmed the blocking in projected coordinates, checking that no tile in one set overlaps a tile in another on the ground.

The second is to score at the rarity the target actually has, and to test only once.
We report the PR-AUC at prevalence ratios of 1:200 to 1:1000 rather than at the 1:5 to 1:20 ratios used for model selection \citep{saito2015precision}, and the recipe and detection settings were scored on the held-out regions exactly once.
The object metrics changed little from validation to test (Section~\ref{sec:testperf}), so the detection settings generalised beyond the data used to choose them.
Reporting at the easier selection prevalence, or tuning against the test set, are the two most common ways a rare-object map can look better on paper than in deployment, and both are avoidable at no cost.

\subsection{Labelling ambiguous pixels}
\label{sec:disc-labelling}
Pixels that cannot be confidently classified are left out of the loss rather than assigned to a class.
An RTS is defined jointly by a barren floor and an active headwall with its cast shadow, although the shadow is not always visible in satellite imagery because of the sun direction and angle.
We nonetheless kept the shadow as a strict requirement, to hold down false positives.
Where an RTS crosses a tile boundary and only one of these features is visible, the object cannot be confirmed without the neighbouring tile, and a partial RTS with only its floor visible is indistinguishable from other bare ground.
Neither remaining class is correct for such pixels, and each choice reveals a specific error: labelling them as RTS treats bare ground as an RTS, and labelling them as background penalises the model for finding a real one.
Assigning them an ignore index avoids both by explicitly stating ignorance rather than guessing.
The argument is one of consistency rather than of measured gain.
We tested the treatment directly and the result was inconclusive (Supplementary Section~S1), so we report the scheme as a design choice available to any rare-feature labelling campaign at negligible cost, not as a demonstrated improvement.

\subsection{Continental-scale inference}
\label{sec:disc-operations}
Three choices from Section~\ref{sec:inference} generalise to any inference run at this scale.
The first is to coordinate workers through shared storage rather than a central scheduler, so the queue is self-balancing, and a run survives interruption without a controlling process, which matters when the fleet is preemptible and spans days (Supplementary Section~S5).
The second is to enforce parity between the settings used to calibrate the model and those used to deploy it.
The temperature, the selection threshold, the numerical precision and the test-time augmentation are read from one configuration and asserted to match at runtime, because a threshold chosen under one precision or augmentation setting is not valid under another.
This closes a subtle path by which a well-validated model can be silently miscalibrated in production.
The third is to rehearse the run before making it.
The failures that stop a continental run lie in the execution path rather than in the model, and they appear only when real cloud infrastructure is exercised.
A staged trial on a few hundred thousand tiles revealed several blocking defects, from container GPU visibility to shared-memory exhaustion under many data-loader workers, none of which was visible in the model's validation record.
It also corrected the throughput estimate, and thus the cost and schedule.
Budgeting an experimental pre-run is as important as budgeting the run itself.

\subsection{Extensibility of the framework}
\label{sec:disc-extensibility}
The framework was built for mapping RTS, but the conditions that shaped it are common in Earth Observation.
RTS are unevenly distributed and heavily outnumbered by background.
They are spread across very different landscapes, and labelling them is slow and costly.
Many other landforms share some or all of these conditions, including other permafrost disturbances such as thermokarst lakes and active-layer detachments.

The most important lesson concerns evaluation, because every other decision rests on it.
A detector for a rare target should be tested the way it will be used.
In practice, this means scoring it on regions it has never seen in training, at the realistic prevalence of the target (Section~\ref{sec:disc-rigour}).
Splitting the data by ecoregion rather than by tile does this across varied landscapes, because it prevents data leakage due to similar environments (Section~\ref{sec:splits}).

An honest evaluation then shows what actually improves a detector.
In this study, more labels helped more than a larger model did, and a foundation model helped only when labels were few (Section~\ref{sec:levers}).
Where labels are scarce, effort is therefore better spent on labels than on model size.
The same evaluation also shows how often a plausible addition fails.
With the full label set, multimodal fusion, foundation-model encoders and self-supervised pretraining all cost far more computation, yet none did better than a convolutional network (Section~\ref{sec:disc-representation}).
A new component should therefore be adopted only if it beats a strong simple baseline under the same protocol.

None of these practices is new on its own.
Other RTS studies have tested models on separate regions \citep{nitze2021testing, lou2026hemispheric}, and have compared architectures and input channels \citep{zeng2026multisource, li2025multimodal, lou2026hemispheric}.
What is new is their combination within one system taken from labelling to a circumpolar map (Section~\ref{sec:disc-comparison}).

Behind these lessons is one design philosophy, shared by the Earth Observation system and the deep learning system.
Each component was kept only when the evidence supported it, so the final system is the simplest one that met the criterion (Section~\ref{sec:experiments}).
New data sources would enter the same way as the channels tested here, through the adoption criterion.
The same discipline shaped the engineering, whose choices apply to any run at this scale (Section~\ref{sec:disc-operations}).

\subsection{Limitations}
\label{sec:disc-limitations}
One limitation concerns recall.
The delivered map recovers about half of the held-out reference objects, at a recall of 0.516, so the detector is deliberately conservative rather than exhaustive.
The choice reflects how the two errors degrade a map.
Missed RTS leave a conservative sample in which density, size and regional patterns remain interpretable.
In contrast, false detections are not distributed at random but concentrate in confusable terrain, so they distort the very patterns the map is read for.
They also cost the reader time, because below some precision the output has to be checked before it can be used, which makes it a candidate list rather than a map.
We judge on these grounds that a map with high precision at an acceptable recall serves more uses than one with high recall at an acceptable precision.
We therefore publish the calibrated probability field, so that a recall-sensitive user can choose a more permissive threshold without any new inference.
Recall at deployment on the 2025 imagery nevertheless remains unmeasured, because no same-year reference polygons exist.
Roughly a quarter of the reference objects carry no detectable signal at any threshold (Section~\ref{sec:testperf}), so recall is bounded by what the imagery shows rather than by where the threshold sits.
These are the features the basemap does not resolve, which the labelling standard of Section~\ref{sec:labels} already excludes.

The second limitation is imagery.
The input is a growing-season optical basemap at a projected resolution of about 4.77\,m, whose ground sample and mosaic composition vary with latitude and acquisition.
Small or spectrally faint RTS near this resolution are the hardest to detect, and the model sees no elevation or radar information that might disambiguate them \citep{rodenhizer2024comparison}.
Elevation is one of these we tested directly, and it added nothing once NDVI was present (Section~\ref{sec:ablations}).
Hence, the limitation lies in the imagery rather than in the choice of auxiliary channel.

The third limitation is domain coverage.
The detector is applied over the region where the 2025 PlanetScope basemap and the permafrost extent overlap, which excludes the far northern permafrost beyond the basemap's guarantee at roughly 74 to 76\,\textdegree{}N, and it represents a single growing season rather than a multi-year state.

\subsection{Future work}
\label{sec:disc-future}
If the limit is the representativeness of the labels rather than the capacity of the model, the most promising extensions are those that add information without demanding new manual annotation.
Hard-negative mining is the leading one, meaning that false positives are identified in the inference output and returned to the training set for the next round.
It is already seeded, because the stratified review of Section~\ref{sec:product} rated 152 of its 280 sampled polygons as incorrect, and those are exactly the confusing background the training set under-samples.
Adding them as negatives would raise precision where it is currently weakest.
The reconstructive form of self-supervised pretraining, which we had considered the clearest such route, was tested and failed (Sections~\ref{sec:pretraining} and~\ref{sec:disc-representation}), so future self-supervision should look to discriminative objectives such as contrastive learning, which do not overwrite the features a strong checkpoint already carries.
Both routes turn spare computation and the inference output itself into supervision, which is cheaper than expanding the label set by hand, where expert disagreement already limits how far manual labelling can scale \citep{nitze2025labeling}.

A second direction is to make detection dynamic.
The deployment demonstrated here is a single-season snapshot.
Yet, much of the scientific value of an RTS inventory lies in change detection, in where RTS initiate and how fast headwalls retreat \citep{lewkowicz2019extremes, barth2025rapid, xia2024widespread}.
Repeat basemaps invite a multi-temporal formulation that detects new and enlarging RTS directly, and pairing detections with elevation change would move towards the volumetric and carbon-flux quantities that motivate RTS mapping in the first place \citep{dai2025volumetric, turetsky2020abrupt}.
Because basemap appearance drifts between years, such an extension will need the model to transfer across acquisitions without relabelling.
This domain adaptation problem \citep{huang2022transferability, lin2025domain} has been partially addressed in this study by interannual train-inference, seeding further investigation.

A third direction concerns coverage and transfer.
The permafrost beyond the basemap's reach, at about 74 to 76\,\textdegree{}N, is unmapped, and reaching it may require fusing sensors with differing resolutions, for which the trained half-resolution capability of Section~\ref{sec:pretraining} serves as a starting point.
Finally, applying the deep learning system to another rare disturbance, such as thermokarst lakes or active-layer detachments, would test how far it generalises.

\section{Conclusions}
\label{sec:conclusions}

We developed and deployed a precision-prioritised deep learning system for detecting retrogressive thaw slumps, a rare target under scarce labels, wall-to-wall across the circumpolar permafrost region.
Four findings carry beyond this study to other rare targets mapped from satellite imagery.

First, model capacity, self-supervised pretraining and an added spectral channel turned out to be substitutes for one another rather than independent gains.
Of 21 candidate channels drawn from four Earth Observation sources, only Sentinel-2 NDVI was adopted, and terrain and satellite embeddings, though each helpful alone, were redundant once NDVI was present.
NDVI gives a convolutional network the same prior that pretraining gives a foundation model, more cheaply, and it adds nothing to a model that already has it.
All three approach a ceiling that the number of labels sets, and none of them raises it.

Second, labels are what limits the system.
The number of positive tiles bounds the detection performance it can reach.
More training tiles mean wider coverage of the variation in RTS form, colour and background landscape, and more examples of what is not an RTS.

Third, evaluation practice determined the credibility of the reported skill.
Ecoregion-blocked splits, prevalence-scaled precision-recall analysis, and a single-shot test protocol produced an honest estimate of deployment performance.

Fourth, deploying the validated detector over 41.6 million tiles covering 20.68 million km\textsuperscript{2} produces a circumpolar map of 19,068 high-confidence RTS detections, with a stratified-sample precision of 0.54 to 0.90.
These are model detections, characterised here by a stratified sample rather than a census-verified inventory.

Taken together, these findings reframe the task.
What limits an operational RTS detector is not the network but the labels, the class imbalance, and how honestly it is evaluated.
RTS mapping is, in this sense, a problem of Earth Observation data and evaluation rather than of network design.
The contribution is to have tested this under one protocol, at the prevalence the target actually has, and to have measured the precision the detector keeps when it is deployed wall-to-wall.

Future work will extend the survey to interannual RTS change detection and to the permafrost beyond the basemap's reach, and will add recall-oriented thresholds for regional studies.
The labelling scheme, adoption criterion, and evaluation protocol are directly reusable for other rare permafrost disturbance features.

\section*{CRediT authorship contribution statement}

\textbf{Yili Yang:} Conceptualization, Methodology, Software, Validation, Formal analysis,
Investigation, Data curation, Writing -- original draft, Writing -- review \& editing, Visualization, Supervision.
\textbf{Heidi Rodenhizer:} Conceptualization, Methodology, Software, Validation, Formal analysis,
Investigation, Data curation, Writing -- review \& editing, Visualization, Supervision.
\textbf{Robb Young:} Software, Data curation.
\textbf{Brendan M. Rogers:} Conceptualization, Writing -- review \& editing, Supervision, Funding acquisition.
\textbf{Marlena Bartkus:} Data curation.
\textbf{Stefano Potter:} Writing -- review \& editing.
\textbf{Greg Fiske:} Writing -- review \& editing, Visualization.
\textbf{Scott Zolkos:} Writing -- review \& editing.
\textbf{Wenwen Li:} Writing -- review \& editing.
\textbf{Chia-Yu Hsu:} Writing -- review \& editing.
\textbf{Luigi Marini:} Resources.
\textbf{Todd C. Nicholson:} Resources.
\textbf{Benjamin C. Maglio:} Writing -- review \& editing.
\textbf{Jennifer Watts:} Resources.
\textbf{Anna Liljedahl:} Resources, Writing -- review \& editing, Funding acquisition.
\textbf{Susan M. Natali:} Writing -- review \& editing, Funding acquisition.

\section*{Declaration of generative AI and AI-assisted technologies in the manuscript preparation process}
During the preparation of this work, the authors used Claude Code for drafting and restructuring text, literature searches (independent DOI verification via Crossref), and generating data visualisations from the underlying research data.
After using this tool, the authors reviewed and edited the content as needed and take full responsibility for the content of the published article.

\section*{Declaration of competing interest}
The authors declare that they have no known competing financial interests or personal relationships that could have appeared to influence the work reported in this paper.

\section*{Data availability}
The code that generates the training data, trains the models, and runs the pan-Arctic inference is openly available at \url{https://github.com/whrc/RTSmapping_v2}, together with the trained model weights.
The refined reference polygons build on the openly available Arctic Retrogressive Thaw Slumps (ARTS) compilation \citep{yang2025arts}.
The PlanetScope Quarterly Basemap imagery cannot be redistributed under its commercial licence, but the deterministic pipeline in the repository reconstructs every training and inference tile exactly from the basemap for any licence holder.
The Sentinel-2 and Google Satellite Embedding inputs are derived from openly available collections through Google Earth Engine.
The raw calibrated probability field and the high-confidence detection map are browsable as provisional model detections through a public Google Earth Engine application at \url{https://abruptthawmapping.projects.earthengine.app/view/south-rts-map}.
The final product is archived in the Arctic Data Center (doi:10.18739/A26970107).

\section*{Acknowledgements}
Funding: This work was supported by the Heising-Simons Foundation, United States [grant number 2021-3040]; Google.org's Impact Challenge for Climate Innovation Program Permafrost Discovery Gateway; and funding catalysed by the TED Audacious Project (Permafrost Pathways).
Computation resources on the Google Cloud Platform were acquired via the Permafrost Discovery Gateway.
The funders had no role in the study design, in the collection, analysis and interpretation of data, in the writing of the report, or in the decision to submit the article for publication.

\clearpage %
\bibliographystyle{elsarticle-harv}
\bibliography{refs}

\end{document}